\documentclass[lettersize,journal]{IEEEtran}
\usepackage{amsmath,amsfonts}
\usepackage{algorithmic}
\usepackage{algorithm}
\usepackage{array}
\usepackage[caption=false,font=normalsize,labelfont=sf,textfont=sf]{subfig}
\usepackage{textcomp}
\usepackage{stfloats}
\usepackage{url}
\usepackage{verbatim}
\usepackage{graphicx}
\usepackage{soul}
\usepackage{pifont}
\usepackage{booktabs}
\usepackage{multirow}
\usepackage[table]{xcolor}
\usepackage{pifont}
\usepackage[most]{tcolorbox}
\usepackage{lipsum}

\usepackage[normalem]{ulem} 

\definecolor{text}{HTML}{E66F51}
\definecolor{image}{HTML}{E9C46B}
\definecolor{video}{HTML}{8AA7DB}
\definecolor{3d}{HTML}{CDC7E5}
\definecolor{multi}{HTML}{00BC7B}
\sethlcolor{blue!3!white} 

\usepackage[colorlinks=true, linkcolor=blue, citecolor=blue]{hyperref}
\begin{document}

\title{
    \raisebox{-1ex}{\includegraphics[width=2cm]{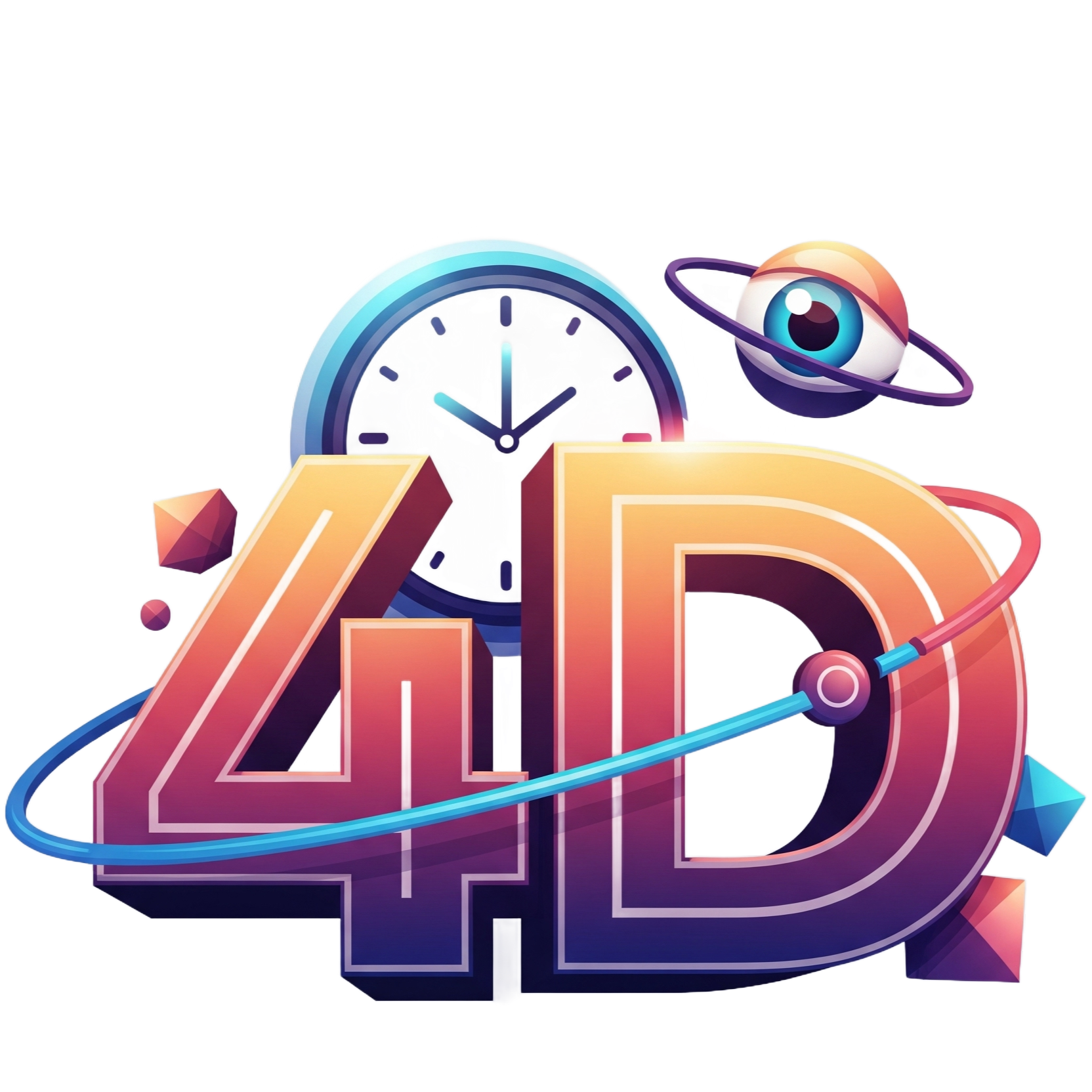}}
    Advances in 4D Generation: Techniques, Challenges, and Future Directions
}

\author{
Qiaowei Miao,~\IEEEmembership{Student Member,~IEEE}, Kehan Li, Jinsheng Quan, Zhiyuan Min, Shaojie Ma, Yichao Xu, Yi Yang,~\IEEEmembership{Senior Member,~IEEE}, Ping Liu$^{*}$, Yawei Luo$^{*}$

\vspace{0.5cm}
\textit{(Survey Paper)}
\vspace{-25pt}

\thanks{
Qiaowei Miao, Kehan Li, Jinsheng Quan, Zhiyuan Min, Shaojie Ma, Yichao Xu, Yi Yang, and Yawei Luo are with the School of Software and Technology, Zhejiang University, China.
Ping Liu is with the Department of Computer Science, University of Nevada Reno. E-mail: QiaoweiMiao@zju.edu.cn; kehan.li@zju.edu.cn; jinshengquan@zju.edu.cn; minzhiyuan@zju.edu.cn; 22351146@zju.edu.cn; yichaoxu@zju.edu.cn; yangyics@zju.edu.cn; pino.pingliu@gmail.com; yaweiluo@zju.edu.cn
}
\thanks{
Corresponding authors: Yawei Luo, Ping Liu.
}
}

\markboth{Journal of \LaTeX\ Class Files,~Vol.~14, No.~8, August~2024}%
{Shell \MakeLowercase{\textit{et al.}}: A Sample Article Using IEEEtran.cls for IEEE Journals}


\maketitle

\begin{abstract}
Generative artificial intelligence has recently progressed from static image and video synthesis to 3D content generation, culminating in the emergence of 4D generation—the task of synthesizing temporally coherent dynamic 3D assets guided by user input. 
As a burgeoning research frontier, 4D generation enables richer interactive and immersive experiences, with applications ranging from digital humans to autonomous driving.
Despite rapid progress, the field lacks a unified understanding of 4D representations, generative frameworks, basic paradigms, and the core technical challenges it faces.
This survey provides a systematic and in-depth review of the 4D generation landscape. 
To comprehensively characterize 4D generation, we first categorize fundamental 4D representations and outline associated techniques for 4D generation. 
We then present an in-depth analysis of representative generative pipelines based on conditions and representation methods. 
Subsequently, we discuss how motion and geometry priors are integrated into 4D outputs to ensure spatio-temporal consistency under various control schemes. 
From an application perspective, this paper summarizes 4D generation tasks in areas such as dynamic object/scene generation, digital human synthesis, editable 4D content, and embodied AI. 
Furthermore, we summarize and multi-dimensionally compare four basic paradigms for 4D generation: End-to-End, Generated-Data-Based, Implicit-Distillation-Based, and Explicit-Supervision-Based.
Concluding our analysis, we highlight five key challenges—consistency, controllability, diversity, efficiency, and fidelity—and contextualize these with current approaches.
By distilling recent advances and outlining open problems, this work offers a comprehensive and forward-looking perspective to guide future research in 4D generation.
The resources compiled in this survey are publicly available at: \href{https://github.com/MiaoQiaowei/Awesome-4D}{https://github.com/MiaoQiaowei/Awesome-4D}.

\end{abstract}


\vspace{-5pt}
\section{Introduction}\label{sec:intro}

In recent years, generative models demonstrate remarkable progress across multiple modalities. Initially centered on static image synthesis~\cite{rombach2022high, saharia2022photorealistic}, research rapidly expands into video generation~\cite{chen2024videocrafter2, chen2023videocrafter1, ho2022video, hoppe2022diffusion, harvey2022flexible}, multiview image generation~\cite{shi2023mvdream, liu2023zero1to3, shi2023zero123++, lin_oneto3d_2024}, and 3D content synthesis~\cite{shi2023mvdream, lin_oneto3d_2024, wang2023score, raj2023dreambooth3d}. 
{Despite these advances, achieving both long-range temporal coherence, such as maintaining consistent object identities across frames, and flexible user control through pose or text guidance remains an open challenge for current 3D generative approaches to create dynamic assets.}

\begin{figure}[!tp]
    \centering
    \includegraphics[width=1\linewidth]{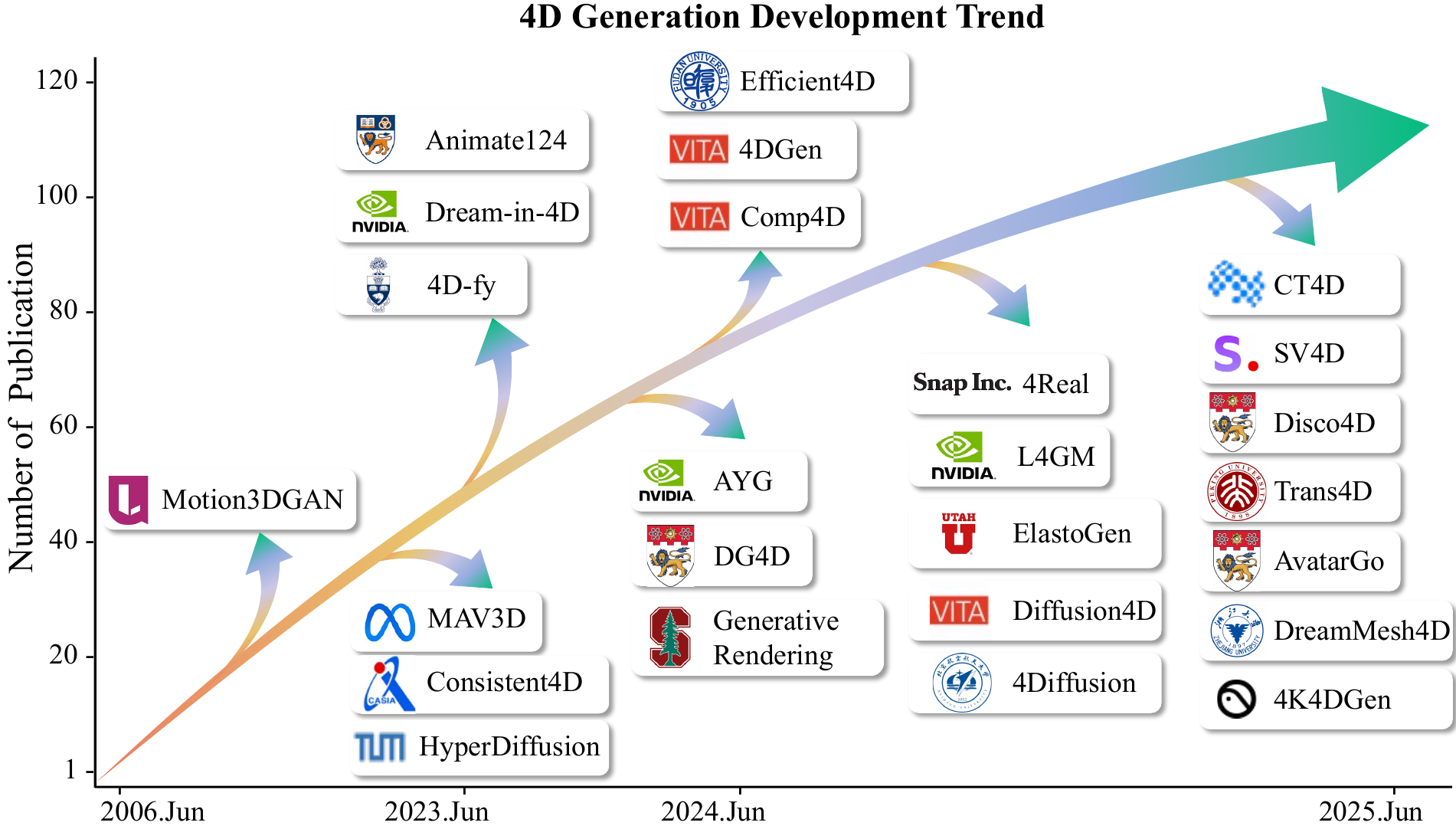}
    \vspace{-20pt}
    \caption{\textbf{4D generation development trends.} The field of 4D generation has grown rapidly, as shown by the sharp increase in annual publications. 
    Representative works from prominent academic and industrial institutions—such as NVIDIA, Snap Inc., TUM, and VITA Lab—highlight the expanding diversity and increasing impact of this emerging research area.}
    \label{fig:trend}
    \vspace{-20pt}
\end{figure}

\begin{figure*}[t]
    \centering
    \includegraphics[width=1\linewidth]{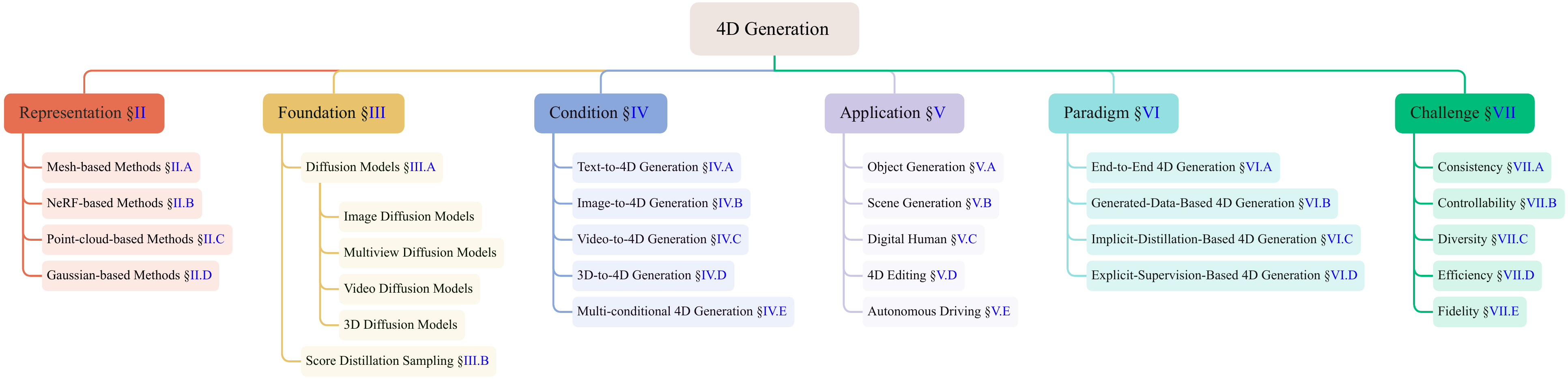}
    \vspace{-20pt}
    \caption{ \textbf{Overall of the survey.} A systematic taxonomy illustrating key aspects of 4D generation, including representation methods, foundational techniques, conditioning strategies, diverse applications, representative paradigms, and core technical challenges. 
    This structured framework clearly delineates current relationships and emerging research directions within the field.
    }
    \vspace{-20pt}
    \label{fig:框架}
\end{figure*}

To overcome these limitations, researchers increasingly turn to 4D generation, a rapidly evolving frontier focusing explicitly on dynamic 3D content synthesis over extended temporal horizons~\cite{singer2023text, bahmani20244dfy, zheng_unified_2024, jiang2023consistent4d, ling2024align, ren2023dreamgaussian4d}.
Specifically, 4D generation refers to the task of synthesizing dynamic 3D content that evolves over time, often with view-consistent geometry and motion. 
By explicitly modeling the temporal dimension, 4D generation unlocks new capabilities for immersive applications in digital humans~\cite{guo2023vid2avatar}, AR/VR~\cite{miao_pla4d_2024}, and autonomous driving~\cite{MagicDrive3D}.  
This trend highlights not only increasing academic attention but also a growing demand for temporally coherent dynamic 3D content in applications such as virtual humans and autonomous systems.

Technically, 4D generation is underpinned by two critical enablers: temporal 3D representations and generative diffusion models.
The first stems from advances in 3D modeling, such as neural radiance fields (NeRFs)~\cite{mildenhall2021nerf}, point-based networks~\cite{qi2017pointnet}, and Gaussian splatting~\cite{kerbl3Dgaussians}, which have been extended into the temporal domain through deformation-aware architectures~\cite{park2021nerfies, wu20244d}. 
These methods model spatial-temporal changes by aligning frame-wise geometry with consistent identities and topologies.
In parallel, generative diffusion models have demonstrated exceptional realism and controllability in image synthesis. 
Recent work~\cite{ren2023dreamgaussian4d, mou2024instruct} has begun to adapt them for 4D content, enabling temporally coherent and semantically editable dynamic assets.

Together, these advances have significantly expanded the design space for 4D generation, supporting a wide range of downstream tasks, including dynamic object~\cite{ling2024align} and scene generation~\cite{yu_4real_2024}, photorealistic digital human modeling~\cite{zhao2022human}, 4D-aware editing~\cite{mou2024instruct}, and interactive simulation for robotics and embodied AI agents~\cite{zhen2025tesseract}. 
This growing demand and expanding set of applications have, in turn, sparked increasing research activity. 
As depicted in Figure~\ref{fig:trend}, annual publications on 4D generation rose from single digits before 2020 to over forty in 2024, underscoring the field’s rapid ascent across computer vision, graphics, and robotics.

Despite its rapid development, the field of 4D generation currently lacks a unified perspective, with existing studies spanning diverse directions.
{To date, there is no comprehensive survey that unifies recent advances, proposes a structured taxonomy, summarizes typical paradigms, or identifies cross-cutting challenges that span representation, generation, and application. }
This lack of a consolidated view hampers both newcomers and established researchers from navigating the field and building upon its foundations.
To address this gap, we present the first systematic survey of 4D generation, with the following contributions:

\begin{itemize}
\item \textbf{Structured synthesis of 4D generation landscape}: We systematically review and organize existing works along three orthogonal axes, including 4D representations (\textit{e.g.}, mesh, point cloud, NeRF), generative mechanisms (\textit{e.g.}, deformation-based, diffusion-based), and application domains (\textit{e.g.}, digital humans, dynamic scenes, embodied agents). 
This three-fold framework provides a holistic view of the field.
\item \textbf{Unified taxonomy and analytical framework}: We propose a unified taxonomy based on the nature of conditioning signals (\textit{e.g.}, text, image, video, 3D, multi-modal), revealing the diverse control strategies in 4D generation. 
Furthermore, we distill four representative generative paradigms: end-to-end, generated-data-based, implicit-distillation-based, and explicit-supervision-based—and introduce a comparative framework for multi-dimensional analysis across supervision, controllability, and temporal consistency.

\item \textbf{Forward-looking challenges and research directions}: We identify key technical challenges, including spatial-temporal consistency, user-controllable generation, computational overhead, and fidelity degradation. 
We also highlight emerging trends such as real-time generation, embodied agent alignment, and efficient training strategies, offering concrete directions for future research.
\end{itemize}


To guide readers through this emerging field, this survey is structured around multiple interrelated components, as illustrated in Figure~\ref{fig:框架}. We begin by reviewing foundational 4D representations, highlighting how static 3D structures are extended into the temporal domain (Section~\ref{sec:related}).
{Subsequently, we introduce technologies highly relevant to 4D generation, drawing comparisons to highlight their distinctions and interrelationships with image, multiview, video, and 3D generation tasks (Section~\ref{sec:tech}).}
Next, we provide an overview of techniques for integrating conditions into the generation process and present a taxonomy classifying generation pipelines based on diverse control modalities, including text, images, videos, 3D data, and various fused controls (Section~\ref{sec:4D}).
Subsequent sections explore practical applications across dynamic objects/scenes, digital humans, and embodied agents (Section~\ref{sec:4Dappl}).
{We then distill existing 4D generation approaches into four key paradigms: End-to-End, Generated-Data-Based, Implicit-Distillation-Based, and Explicit-Supervision-Based (Section~\ref{sec:paradigms}). We further analyze the key technical challenges, such as consistency, controllability, diversity, efficiency, and fidelity (Section~\ref{sec:4Dcha}).}
Finally, we conclude with a discussion of emerging trends and prospective research directions that may shape the future of 4D generation (Section~\ref{sec:discussion}).
Through this structure, our survey offers a cohesive roadmap of the 4D generation landscape, uniquely integrating a structured taxonomy, a comprehensive synthesis of emerging generative techniques, a distillation of paradigms, and an insightful analysis of open challenges to guide future research.

\vspace{-10pt}
\section{4D REPRESENTATION}\label{sec:related}

\begin{figure}[tp]
    \centering
    \includegraphics[width=1\linewidth]{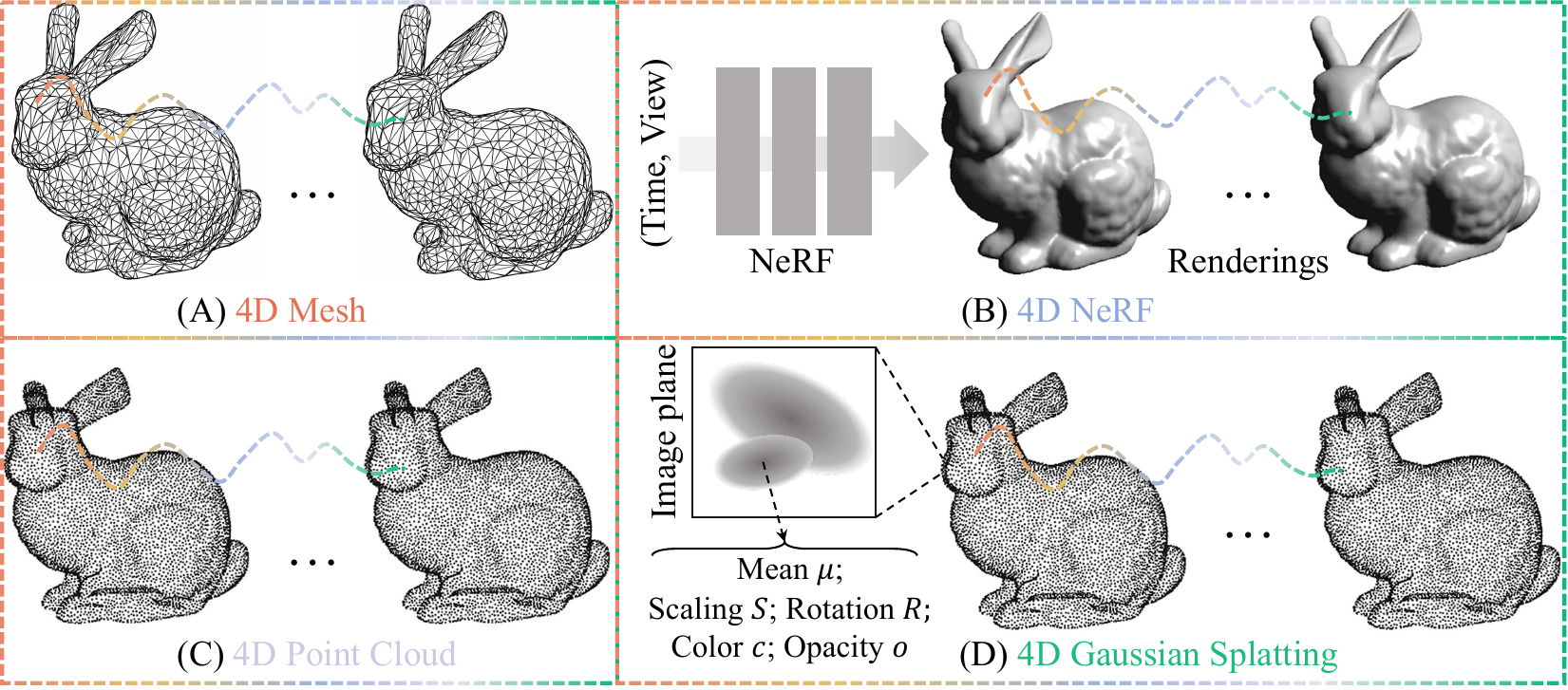}
    \vspace{-15pt}
    \caption{\textbf{Illustration of four major 4D representation methods}:
(A) \textcolor{text!100}{4D Mesh} explicitly represents evolving surfaces by continuously updating vertex positions and connectivity;
(B) \textcolor{video!100}{4D NeRF} implicitly encodes scenes as continuous volumetric functions, providing smooth renderings across views and time;
(C) \textcolor{3d!100}{4D Point Cloud} represents dynamics via discrete point displacements, offering flexibility and simplicity;
(D) \textcolor{multi!100}{4D Gaussian Splatting} employs anisotropic Gaussian primitives for compact, high-quality representation and rendering.
While 4D NeRF employs an implicit representation approach, the other three methods explicitly encode geometric entities.}
    \label{fig:4d}
    \vspace{-15pt}
\end{figure}

In computer vision and graphics, effectively modeling dynamic real-world objects requires representations that explicitly integrate temporal information. 
Traditional 3D representations focus primarily on static shape and appearance, yet struggle to accurately capture continuous geometric transformations, complex topology changes, and appearance variations across time. 
To address these limitations, 4D representations incorporate the temporal dimension into existing 3D models, enabling accurate and efficient rendering of dynamic, high-resolution scenes.
In Figure~\ref{fig:4d}, we summarize four prevalent classes of 4D representations, each capturing temporal variations through distinct modeling paradigms. 
These representations are reviewed in detail in the following subsections.

\vspace{-16pt}
\subsection{4D Mesh}
\vspace{-16pt}
4D mesh representation explicitly captures temporal dynamics by continuously updating mesh vertex positions and, in some cases, adapting mesh topology. This capability makes them highly suitable for representing evolving surfaces, particularly in character animation and human motion capture. For example, the Skinned Multi-Person Linear Model (SMPL)~\cite{loper2023smpl} parameterizes the human body as a triangular mesh, employing linear blend skinning techniques to generate realistic dynamic human movements. Extensions such as STAR~\cite{osman2020star} further enhance pose reconstruction accuracy through sparse optimization, while GHUM~\cite{xu2020ghum} captures diverse body shapes and poses simultaneously, significantly improving marker-less dynamic modeling quality. 

Moving beyond character animation, dynamic meshes integrated with physical simulations effectively model complex phenomena such as soft-body dynamics and fluid interactions. Traditional simulation approaches, such as finite element methods~\cite{reddy1993introduction}, capture key physical properties including elasticity, rigidity, and viscosity, enabling effective applications in medical and engineering simulations. However, these approaches face substantial computational challenges for large-scale deformations and intricate interactions. 
Recent deep-learning-based techniques, such as {Pfaff \textit{et al.}~\cite{pfaff2020learning} and Sanchez-Gonzalez \textit{et al.}~\cite{sanchez2020learning}}, employ graph neural networks and mesh-based architectures to improve computational efficiency significantly. 
To further enhance realism and enable detailed control, recent approaches integrate differentiable rendering into dynamic mesh representations. Differentiable rendering allows end-to-end optimization of geometry, material, and lighting parameters through image-based supervision, reducing reliance on manual tuning. Methods such as DreamMesh4D~\cite{li_dreammesh4d_2024} integrate explicit deformation modeling with neural rendering frameworks, significantly improving dynamic scene quality. However, dynamic meshes often struggle with intricate topology changes and volumetric phenomena, motivating a shift toward implicit representations like Neural Radiance Fields.

\subsection{4D NeRF}
Unlike explicit representations such as dynamic meshes, which directly encode vertex positions and topology, implicit methods like Neural Radiance Fields (NeRF) model scenes as continuous volumetric functions, naturally handling complex topology changes and volumetric details. Extending NeRF to the temporal domain has led to a variety of 4D NeRF methods, each addressing different aspects of dynamic scene modeling.

4D NeRF approaches can be broadly categorized into temporal latent representations, deformation-based methods, and scene-flow-based techniques. One of the earliest efforts to adapt NeRF for 4D tasks is Neural 3D Video~\cite{li2022neural}, which introduces temporal latent variables for efficient rendering of 4D scenes. Utilizing hierarchical training and importance sampling, it achieves scalable performance, with the DyNeRF dataset serving as a benchmark for dynamic NeRF modeling. 
Another direction explores deformation fields, as demonstrated by Nerfies~\cite{park2021nerfies} and D-Nerf~\cite{pumarola2021d}, which extend NeRF to 4D by integrating temporal variables into deformation fields.
{Nerfies~\cite{park2021nerfies} decomposes 4D space into a static NeRF-based template and time-dependent deformation fields; D-Nerf~\cite{pumarola2021d}, on the other hand, predicts spatial displacements over time to model dynamic scenes.} 
Scene-flow-based methods, such as Neural Scene Flow Fields~\cite{li2021neural}, jointly encode geometry and motion dynamics by training neural fields to predict scene flow, effectively modeling complex temporal changes.

Despite these successes, existing 4D NeRF methods face significant challenges regarding computational efficiency and scalability due to their high-dimensional representations. To address these issues, recent research explores spatial decomposition strategies to reduce complexity. One group of methods, including HyperReel~\cite{attal2023hyperreel} and NeRFPlayer~\cite{song2023nerfplayer}, leverages keyframe sampling and region-based decomposition to efficiently represent dynamic scenes. Another class of methods, such as HexPlane~\cite{cao2023hexplane}, K-Planes~\cite{fridovich2023k}, and Tensor4D~\cite{shao2023tensor4d}, factorizes the 4D space-time representation into compact planar features, achieving substantial computational savings and improved representation expressiveness. Nevertheless, the computational complexity of implicit representations such as NeRF still poses significant limitations, particularly impacting their performance in real-time rendering and interactive editing scenarios. To mitigate these limitations, recent studies have started exploring alternative methods that prioritize lightweight structures and explicit representations, offering promising complementary solutions.

\vspace{-10pt}
\subsection{4D Point Cloud}
Point clouds represent three-dimensional shapes as unstructured sets of discrete points~\cite{qi2017pointnet}. Unlike mesh-based methods with explicit geometric representations or implicit volumetric methods like NeRF, point clouds offer greater flexibility, particularly in efficiently capturing complex and dynamically changing topologies. Their lightweight and sparse representation naturally facilitates real-time rendering, direct manipulation, and scalable reconstruction of dynamic scenes.

Early research in dynamic point cloud reconstruction primarily relied on multiview geometry to reconstruct dynamic scenes from multiple camera views. For instance, Mustafa \textit{et al.}~\cite{mustafa2015general, mustafa2016temporally} introduced a 4D reconstruction framework leveraging wide-baseline camera setups, effectively capturing dynamic events in controlled settings.  Similarly, Wand \textit{et al.}~\cite{wand2007reconstruction} proposed methods for reconstructing dynamic 3D geometries directly from point clouds, achieving compelling results for smoothly deforming surfaces. Although these multiview-based approaches achieve good accuracy under ideal conditions, they typically require dense temporal sampling and extensive camera coverage, limiting their practical scalability and applicability in unconstrained real-world scenarios.

To overcome these limitations, researchers have explored template-guided reconstruction strategies, where predefined geometric templates impose effective constraints, significantly reducing ambiguities inherent in multiview reconstruction~\cite{alldieck2018video, dong20174d, kanazawa2019learning, zheng20174d, tung2017self}. These methods have substantially improved reconstruction accuracy, particularly for structured and domain-specific objects like human bodies, faces, and hands. However, reliance on high-quality, category-specific templates~\cite{blanz2023morphable, loper2023smpl, pishchulin2017building, romero2022embodied} restricts their applicability to general dynamic scenes and diverse shape variations, highlighting the need for more generalized, adaptable, and efficient real-time approaches. These challenges motivate recent efforts toward exploring flexible yet efficient 4D representations, such as Gaussian Splatting, offering compact modeling, rapid rendering, and enhanced adaptability across dynamic scenarios, as discussed next.

\vspace{-10pt}
\subsection{4D Gaussian Splatting}
While dynamic point cloud methods flexibly represent complex spatiotemporal variations, they typically face challenges such as sparsity, limited semantic coherence, and computational complexity, making it difficult to achieve high-quality continuous rendering. To overcome these limitations, recent methods have introduced 4D Gaussian Splatting (4D GS), extending static Gaussian Splatting by incorporating temporal dimensions. By encoding spatial and temporal dynamics using anisotropic Gaussian primitives and deformation fields, 4D GS efficiently models object motions and scene dynamics, significantly improving rendering quality and efficiency~\cite{wu2024recent}.

Recent 4D GS approaches~\cite{yang2024deformable, wu20244d, zhou2024drivinggaussian, huang2024sc, lin2024gaussian, li2024spacetime, xie2024physgaussian, kratimenos2023dynmf, you2024decoupling, lu2024manigaussian, luiten2023dynamic, guo2024motion} typically fall into iterative or deformation-based methods. Iterative methods, such as D-3DGS~\cite{luiten2023dynamic}, independently optimize Gaussian parameters frame-by-frame, enforcing temporal coherence through inter-frame constraints. While interpretable, iterative methods typically face challenges with occlusions and unseen regions, often benefiting from more comprehensive camera coverage. In contrast, deformation-based methods~\cite{yang2024deformable, wu20244d} adopt a shared canonical representation with deformation nets to efficiently predict temporal Gaussian offsets, improving scalability, robustness, and computational efficiency, though accurate deformation modeling remains essential.
{Gaussian Splatting-based methods exhibit strong potential in scalability, motion interpretability, and temporal coherence. 
Future improvements may focus on better handling of novel object emergence and monocular video reconstruction, which remain active areas of research.}
\ding{42} \hl{Among the four primary 4D representations, 4D Mesh, 4D Point Cloud, and 4D Gaussian Splatting are explicit methods, inherently incorporating discernible geometric entities for dynamic modeling. These excel in interpretability and direct manipulation, making them ideal for applications requiring fine control and real-time interaction.
In contrast, 4D NeRF implicitly represents 4D content via learned neural networks, excelling at capturing complex volumetric and topological changes despite often having high computational demands.
Specifically, 4D Mesh adjusts connectivity, 4D Point Clouds track point displacements, and 4D Gaussian Splatting controls primitive displacement, rotation, and scale for compact, expressive dynamic modeling. These diverse representations collectively provide a comprehensive foundation for dynamic 4D content generation, each suited to specific application contexts and technical requirements.}


\section{Foundational Techniques for 4D Generation}\label{sec:tech}

\begin{figure}[tp]
    \centering
    \includegraphics[width=1\linewidth]{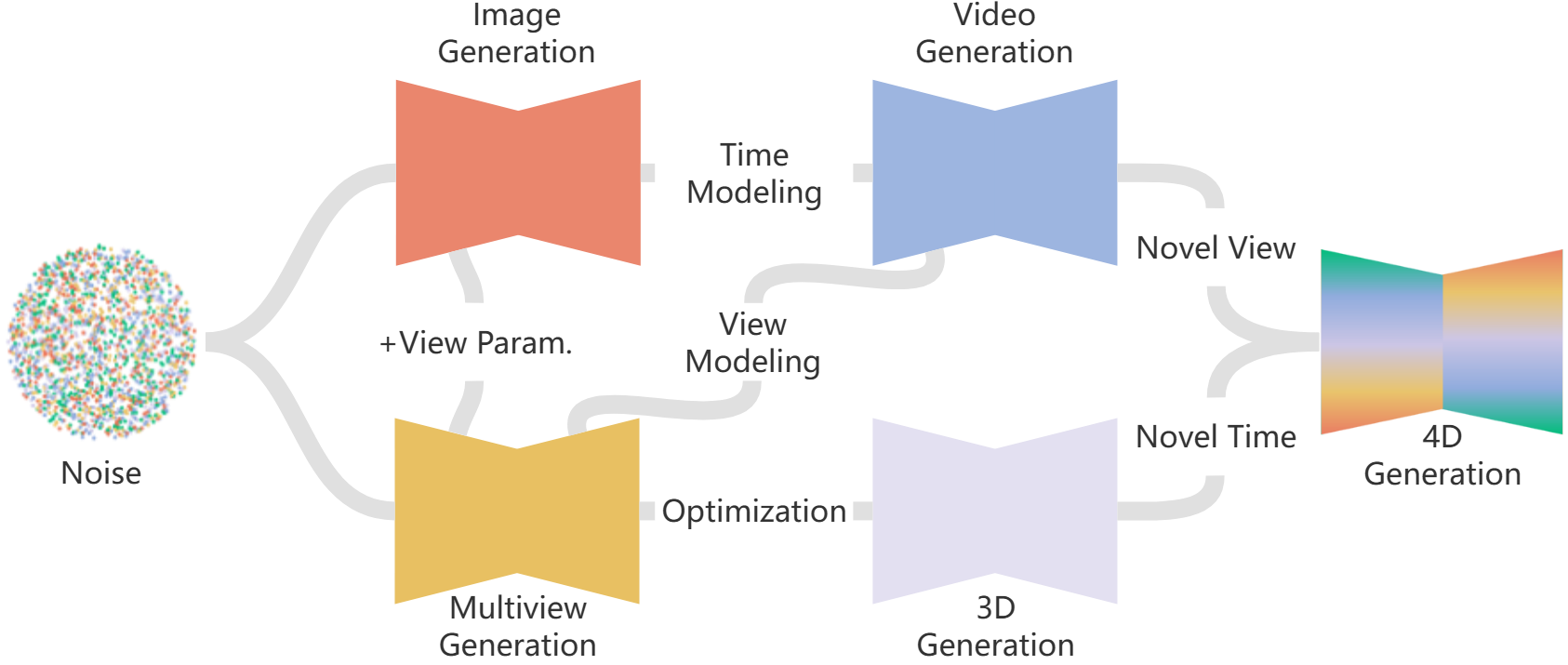}
    \vspace{-20pt}
    \caption{\textbf{4D generation vs. related generative paradigms.} Image generation methodologies focus on synthesizing static, single-viewpoint imagery. Video generation incorporates temporal dynamics, typically focusing on content viewed primarily from a single or constrained set of viewpoints. 
    3D generation synthesizes static geometric models, while multiview generation simultaneously produces images across multiple viewpoints at a single time instance. 
    Distinctively, 4D generation synthesizes dynamic assets that integrate both spatio-temporal coherence and multiview observability.}
    \label{fig:first}
    \vspace{-16pt}
\end{figure}

Building upon the 4D representations introduced earlier, foundational generative techniques such as diffusion models and Score Distillation Sampling (SDS) have become pivotal for addressing core challenges in 4D generation.
This section reviews two foundational methods central to recent advances in 4D generation: diffusion models~\cite{rombach2022high,blattmann2023stable}, renowned for their ability to synthesize realistic dynamic content; and score distillation sampling~\cite{poole2022dreamfusion}, specifically designed to optimize implicit neural representations.
The underlying principles, recent advancements, practical applications, and unique strengths and limitations of each method will be systematically discussed in the following subsections.

\vspace{-10pt}
\subsection{Diffusion Models}
As foundational generative methods, diffusion models~\cite{ho2020denoising, dhariwal2021diffusion, song2020score} have recently emerged as central techniques for generating high-quality content due to their flexibility and visual fidelity. 
These models iteratively remove Gaussian noise from noisy inputs via neural network-based denoising, achieving exceptional generative results. 
{
Specifically, they involve two core processes: a forward diffusion process that systematically adds noise to data, and a reverse denoising process that iteratively reconstructs the original data.
}

\noindent\textit{Forward Diffusion Process.} Gaussian noise is gradually added to the data through a predefined Markov chain~\cite{norris1998markov,rombach2022high}, converting original clean data into pure noise. 
This process is formulated as:
\begin{equation}
q(x_t|x_{t-1}) = \mathcal{N}(x_t; \sqrt{1-\beta_t}x_{t-1}, \beta_t\mathit{I}),
\end{equation}
where $\beta_t$ is a variance schedule that controls the amount of noise added at step $t$.
The direct sampling of any noisy version $x_t$ given original data $x_0$  is efficiently performed by:
\begin{equation}
q(x_t|x_0) = \mathcal{N}(x_t; \sqrt{\bar{\alpha}_t}x_0, (1-\bar{\alpha}_t)\mathit{I}).
\end{equation}

\noindent \textit{Reverse Denoising Process.} The reverse process reconstructs the original data from pure noise, defined by:
\begin{equation}
p_\theta(x_{0:T}) = p(x_T)\prod_{t=1}^{T}p_\theta(x_{t-1}|x_t),
\end{equation}
where a neural network predicts the noise at each step. The training objective minimizes the difference between predicted and actual noise:
\begin{equation}
    \mathbb{E}_{t \sim [1, T], x_0 \sim q(x_0), \epsilon \sim \mathcal{N}(0, \mathit{I})} \left[ \lambda(t) \|\epsilon - \epsilon_\theta(x_t, t)\|^2 \right],
\end{equation}
where $\lambda(t)$ is a weighting function that emphasizes different steps during optimization.
Based on these two core processes, diffusion models have been successfully adapted and extended to various domains, including image~\cite{rombach2022high,zhang2023adding,ruiz2023dreambooth,yu2024metaearth}, multiview~\cite{shi2023mvdream,liu2023zero1to3,shi2023zero123++,lin_oneto3d_2024}, video~\cite{sora,ho2022video,chen2023videocrafter1,chen2024videocrafter2,blattmann2023stable,yuan2025magictime}, and 3D~\cite{chen2023scenedreamer,cao2025difftf++,xue2025gen}.

Diffusion models have become foundational to 4D generation due to their exceptional capability in synthesizing high-quality and coherent dynamic content.
Advances from related domains—such as image diffusion models (providing spatial realism), multiview diffusion models (ensuring cross-view consistency), video diffusion models (capturing temporal coherence), and 3D diffusion models (establishing volumetric and geometric fidelity)—collectively support the multifaceted requirements of spatiotemporally consistent 4D generation tasks. 
As illustrated in Figure~\ref{fig:first}, 4D generation uniquely demands coherence across spatial, temporal, and viewpoint dimensions, setting it apart from conventional image, multiview, video, and 3D generation tasks.
In particular, unlike multiview video synthesis~\cite{li2024vivid} or purely rigging-based approaches for driving 3D targets~\cite{Guo_2025_CVPR}, 4D generation enables the rendering of dynamic scenes that can be freely explored across both time and viewpoint axes, rather than being constrained to fixed perspectives, nor does the motion state of the 4D output rely on manual control.

\vspace{-10pt}
\subsection{Score Distillation Sampling}

{
Building upon the foundational principles of diffusion models discussed previously, Score Distillation Sampling (SDS)~\cite{poole2022dreamfusion, liu2023zero1to3,shi2023mvdream,bahmani20244dfy,miao_pla4d_2024} offers a powerful optimization strategy, widely utilized in diffusion-based 4D generation, specifically designed to leverage diffusion models’ generative priors for enhancing 4D generation results.
}
The central idea of SDS is aligning rendered image features with prior knowledge encoded in the diffusion model, ensuring accurate noise prediction by the diffusion network. Formally, a 4D model parameterized by $\theta$ renders an image $x = g(\theta)$ at specific camera poses, with rendering function $g$. A diffusion model, typically employing a UNet as a denoising network $\epsilon_{\phi}(x;y,t)$, predicts noise added at timestep $t$ given control conditions $y$ (such as text, skeleton, or lighting). SDS optimizes the 4D model parameters $\theta$ by computing the gradient of the SDS loss as follows:
\begin{align}
    \nabla_{\theta}\mathcal{L}_{SDS} = \mathbb{E}_{t,\epsilon}\left[w(t)(\epsilon_{\phi}(x;y,t)-\epsilon)\frac{\partial x}{\partial \theta}\right],
\end{align}
where $w(t)$ is a timestep-dependent weighting function. Through iterative optimization, SDS effectively aligns the rendered image with diffusion priors, indirectly enhancing the geometry and temporal coherence of the underlying 4D representations. The SDS framework has become crucial for advancing structural fidelity and dynamic consistency in recent 4D generative modeling, with ongoing research targeting improved computational efficiency and generalization.

\section{Method and Taxonomy for 4D Generation } \label{sec:4D}

\begin{figure*}[ht]
    \centering
    \includegraphics[width=1\linewidth]{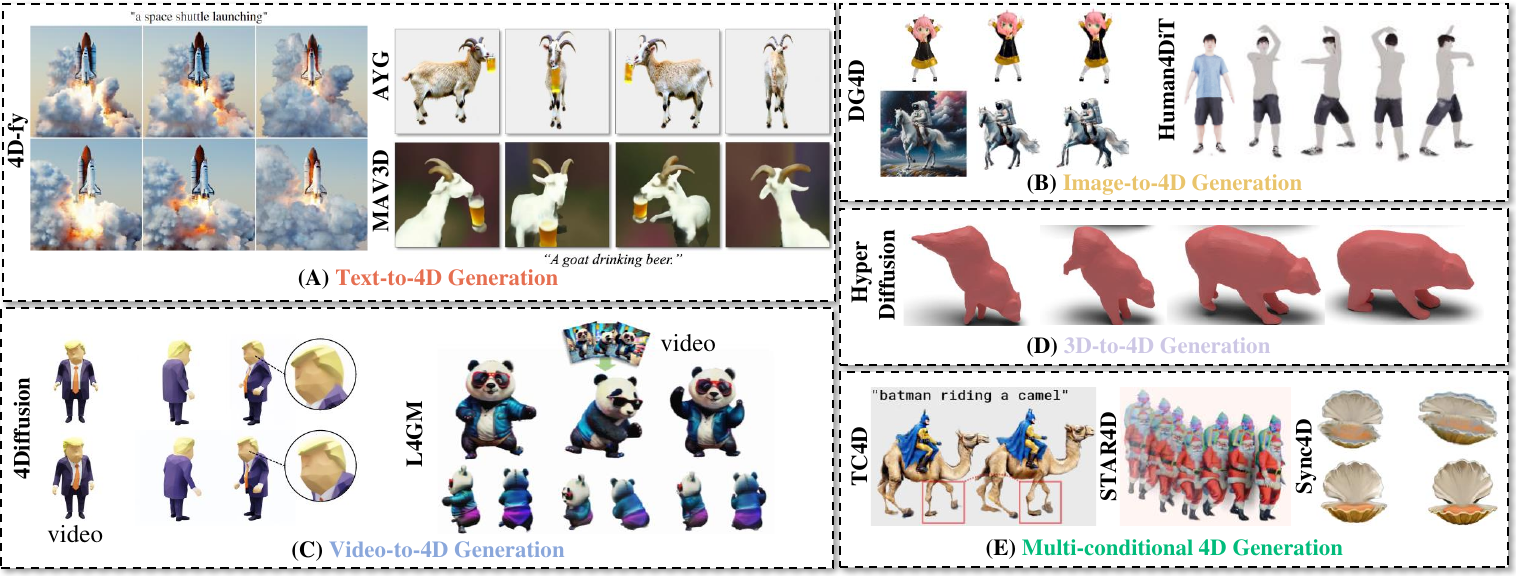}
    \vspace{-18pt}
    \caption{
    \textbf{Representative directions in 4D generation.} Based on different control modalities, 4D generation tasks are categorized into five key domains:
(A) \textcolor{text!100}{Text-to-4D Generation}, where methods such as 4D-fy~\cite{bahmani20244dfy}, MAV3D~\cite{singer2023text}, and AYG~\cite{ling2024align} enable the generation of diverse 4D assets using text as the control condition;
(B) \textcolor{image!100}{Image-to-4D Generation}, exemplified by DreamGaussian4D~\cite{ren2023dreamgaussian4d} (DG4D) and Human4DiT~\cite{shao_human4dit_2024}, which focuses on faithfully reconstructing 4D assets from input images;
(C) \textcolor{video!100}{Video-to-4D Generation}, as demonstrated by 4Diffusion~\cite{zhang_4diffusion_2024} and L4GM~\cite{ren2024l4gm}, emphasizes maintaining spatial consistency over time in generated 4D sequences;
(D) \textcolor{3d!100}{3D-to-4D Generation}, like HyperDiffusion~\cite{dou_dynamic_2024}, extends static 3D assets into the temporal dimension to create dynamic 4D outputs;
(E) \textcolor{multi!100}{Multi-conditional 4D Generation}, showcased by TC4D~\cite{bahmani_tc4d_2024}, STAR4D~\cite{chai_star_2024} and Sync4D~\cite{fu_sync4d_2024}, integrates multiple control conditions to achieve precise and controllable 4D generation.}
    \label{fig:control}
    \vspace{-20pt}
\end{figure*}

\begin{table*}[th!]
\caption{\textbf{A summary of representative methods in 4D generation categorized by input conditions}, including text, image, video, 3D, and multimodal inputs.  
We further summarize these into four distinct paradigms: End-to-End (E2E), Generated-Data-Based (GDB), Implicit-Distillation-Based (IDB), and Explicit-Supervision-Based (ESB) 4D generation. 
Each method is further characterized by its representation type, optimization approach, and primary research motivations such as consistency, controllability, diversity, efficiency, and fidelity.}
\vspace{-8pt}
\centering
\resizebox{\textwidth}{!}{
\begin{tabular}{lccccc} 
\toprule
\textbf{Methods}  & \textbf{Representation} & \textbf{Optimization} & \textbf{Paradigm}  & \textbf{Condition} & \textbf{Main Motivation} \\

\midrule
\rowcolor{text!20}
STAR~\cite{chai_star_2024} & 3D model+Deformation net  & SDS & IDB  & Text  & Consistency\&Diversity\&Fidelity  \\ \rowcolor{text!20}
Dream-in-4D~\cite{zheng_unified_2024}  & NeRF  & SDS & IDB  & Text & Consistency\&Fidelity  \\ \rowcolor{text!20}
MAV3D~\cite{singer2023text}  & NeRF  & SDS & IDB  & Text  & Consistency  \\ \rowcolor{text!20}
4D-fy~\cite{bahmani20244dfy} & NeRF  & SDS & IDB & Text  & Fidelity  \\ \rowcolor{text!20}
CT4D~\cite{chen_ct4d_2024} & NeRF+Mesh  & SDS+MSE & IDB+ESB & Text  & Consistency  \\ \rowcolor{text!20}
AYG~\cite{ling2024align} & Gaussian primitive  & SDS & IDB  & Text  & Consistency  \\ \rowcolor{text!20}
Trans4D~\cite{zeng_trans4d_2024} & Gaussian primitive  & SDS & IDB  & Text  & Efficiency  \\ \rowcolor{text!20}
Comp4D~\cite{xu_comp4d_2024} & Gaussian primitive  & SDS & IDB  & Text  & Controllability\&Fidelity  \\ \rowcolor{text!20}
AvatarGO~\cite{cao_avatargo_2024} & Gaussian primitive  & SDS & IDB  & Text  & Fidelity  \\ \rowcolor{text!20}
4Real~\cite{yu_4real_2024} & Gaussian primitive  & SDS+MSE & IDB+ESB & Text  & Consistency\&Efficiency  \\ \rowcolor{text!20}
AvatarCLIP~\cite{hong2022avatarclip}  &  Mesh  &  CLIP-guided loss & ESB &  Text  &  Consistency\&Fidelity  \\ \rowcolor{text!20}
GAvatar  &  Gaussian primitive+Mesh  &  SDS & IDB  &  Text  &  Consistency\&Efficiency\&Fidelity  \\\rowcolor{text!20}
PLA4D~\cite{miao_pla4d_2024} & Gaussian primitive+Mesh  & SDS+MSE & IDB+ESB & Text  & Consistency\&Controllability\&Diversity\&Efficiency\&Fidelity   \\ \rowcolor{text!20}
STP4D~\cite{deng2025stp4d} & Gaussian primitive  & Diffusion loss & E2E & Text  & Consistency\&Efficiency  \\ \rowcolor{text!20}
TextMesh4D~\cite{TextMesh4D} & Mesh  & SDS+L2  & IDB & Text  & Consistency\&Fidelity   \\ \rowcolor{text!20}
Instruct 4D-to-4D~\cite{mou2024instruct} & NeRF & MSE &GDB & Text  & Consistency\&Controllability\&Fidelity  \\ \rowcolor{text!20}
Control4D~\cite{shao2024control4d} & NeRF  & GAN loss &ESB & Text  & Controllability\&Diversity  \\ \rowcolor{image!20}

\midrule
DKG4D~\cite{song_automatic_2023}  & VAE  & MSE & ESB & Image & Consistency\&Controllability  \\ \rowcolor{image!20}
Diffusion$^{2}$~\cite{yang_diffusionmbox2_2024} & Gaussian primitive  & MSE & ESB & Image  & Consistency\&Efficiency\&Fidelity \\ \rowcolor{image!20}
Disco4D~\cite{pang_disco4d_2024} & Gaussian primitive  & SDS+MSE & IDB+ESB  & Image  & Consistency\&Fidelity \\ \rowcolor{image!20}
DG4D~\cite{ren2023dreamgaussian4d} & Gaussian primitive  & SDS+MSE & IDB+ESB & Image  & Consistency\&Controllability\&Diversity  \\ \rowcolor{image!20}
TwoSquared~\cite{sang2025twosquared}& Mesh  & L1+L2 & ESB  & Image & Consistency \\ \rowcolor{image!20}
4K4DGen~\cite{li_4k4dgen_2024} & Gaussian primitive  & L1  & ESB & Panorama  & Consistency \\ \rowcolor{image!20}
HoloTime~\cite{zhou2025holotime}& Gaussian primitive  & L1  & ESB & Panorama & Consistency\&Fidelity \\ \rowcolor{video!20}

\midrule
DiST-4D~\cite{guo2025dist}& Video  & MSE+v-prediction loss & ESB  & Video & Consistency\&Fidelity \\ \rowcolor{video!20}
Zero4D~\cite{park2025zero4d} & Video  & - & EDB   & video & Consistency\&Fidelity \\ \rowcolor{video!20}
Consistent4D~\cite{jiang2023consistent4d} & NeRF  & SDS & IDB & Video  & Consistency  \\ \rowcolor{video!20}
DS4D~\cite{yang2025not}& NeRF  & SDS  & IDB  & Video & Consistency\&Fidelity \\ \rowcolor{video!20}
SV4D~\cite{xie_sv4d_2024} & NeRF  & MSE  & GDB  & Video  & Consistency \\ \rowcolor{video!20}
4Diffusion~\cite{zhang_4diffusion_2024} & NeRF  & SDS & IDB  & Video  & Consistency\&Fidelity  \\ \rowcolor{video!20}
SC4D~\cite{wu_sc4d_2024} & Gaussian primitive  & SDS & IDB & Video  & Consistency\&Efficiency\&Fidelity \\ \rowcolor{video!20}
4D-Editor~\cite{jiang20234d} & NeRF  & MSE  &ESB & Video  & Consistency \\\rowcolor{video!20}
L4GM~\cite{ren2024l4gm} & Gaussian primitive  & MSE & E2E  & Video  & Consistency\\ \rowcolor{video!20}
Efficient4D~\cite{pan2024efficient4d} & Gaussian primitive  & SDS+MSE & GDB  & Video & Consistency\&Efficiency \\ \rowcolor{video!20}
Vidu4D~\cite{wang2024vidu4d} & Gaussian primitive  &  Normal consistency loss  & ESB & Video  & Consistency  \\ \rowcolor{video!20}
AR4D~\cite{zhu2025ar4d} & Gaussian primitive  & L1 & ESB   & Video  & Efficiency  \\ \rowcolor{video!20}
MVTokenFlow~\cite{huang2025mvtokenflow} & Gaussian primitive  & L1 & ESB  & Video  & Consistency  \\ \rowcolor{video!20}
STAG4D~\cite{zeng2024stag4d} & Gaussian primitive  & SDS+MSE & IDB+ESB   & Video  & Consistency\&Diversity  \\ \rowcolor{video!20}
In-2-4D~\cite{nag20252} & Gaussian primitive  & L1 & ESB  & Video  & Consistency\&Fidelity  \\ \rowcolor{video!20}
FB-4D~\cite{li2025fb}& Gaussian primitive  & SDS & IDB  & video & Consistency\&Fidelity \\ \rowcolor{video!20}
Video4DGen~\cite{wang2025video4dgen}& Gaussian primitive  & L2 & ESB  & Video & Fidelity \\ \rowcolor{video!20}
CAT4D~\cite{wu2025cat4d} & Gaussian primitive  & L1+DSSIM+LPIPS & IDB & Video & Consistency\&Fidelity \\ \rowcolor{video!20}
DreamMesh4D~\cite{li_dreammesh4d_2024} & Gaussian primitive+Mesh  & SDS+MSE & IDB+ESB & Video  & Efficiency\&Controllability \\ \rowcolor{3d!20}

\midrule
Motion3DGAN~\cite{otberdout_3d_2021} & Mesh & L1 & ESB  & Mesh & Diversity\\ \rowcolor{3d!20} 
HyperDiffusion~\cite{erkocc2023hyperdiffusion} & Implicit neural field  & BCE & ESB & MLPs & Fidelity \\ \rowcolor{3d!20} 
ElastoGen~\cite{feng2024elastogen} & Grid  & Fitting error & ESB & Physical parameters & Consistency \\ \rowcolor{3d!20}
ISA-DiT~\cite{shao2025interspatial}& Video  & L1+GAN+KL & ESB  & SMPL  & Consistency \\ \rowcolor{multi!20}
\midrule

Animate124~\cite{zhao_animate124_2023} & NeRF  & SDS+MSE & IDB+ESB  & Text+Image  & Fidelity  \\ \rowcolor{multi!20}
Beyond Skeletons~\cite{yang_beyond_2024} & Mesh  & L1 & IDB   & Text+Image  & Consistency\&Controllability\&Fidelity  \\ \rowcolor{multi!20}
EG4D~\cite{sun_eg4d_2024} & Gaussian primitive  & L1 & IDB  & Text+Image & Consistency \\ \rowcolor{multi!20}
Animate3D~\cite{jiang2024animate3d} & Gaussian primitive  & SDS & IDB & Text+3D & Consistency\&Controllability \\ \rowcolor{multi!20}
Phy124~\cite{lin_phy124_2024} & Gaussian primitive+Material point  & SDS & IDB    & Text+Image  & Consistency\&Controllability\&Efficiency\&Fidelity \\ \rowcolor{multi!20}
Generative Rendering~\cite{cai_generative_2024} & Video  & -  & - & Text+Mesh  & Consistency\&Controllability  \\ \rowcolor{multi!20}
AnimateAnyMesh~\cite{wu2025animateanymesh} & Mesh  & KL loss & E2E & Text+Mesh &Consistency\&Controllability\&Efficiency\&Fidelity\\ \rowcolor{multi!20}
Sketch-2-4D~\cite{yang_beyond_2024} & NeRF  & SDS & IDB  & Text+Sketch  & Consistency\&Controllability  \\ \rowcolor{multi!20}
TC4D~\cite{bahmani_tc4d_2024}  & NeRF  & SDS+MSE & IDB+ESB  & Text+Traj. & Consistency\&Controllability  \\ \rowcolor{multi!20}
DriveDreamer~\cite{wang2024drivedreamer}  & Videos & L1+MSE & E2E   & Text+Traffic constraints/Actions  & Controllability\\ \rowcolor{multi!20}
Diffusion4D~\cite{liang_diffusion4d_2024} & Gaussian primitive  & L2 & ESB  & Text/Image  & Consistency\&Efficiency  \\ \rowcolor{multi!20}
4DGen~\cite{yin_4dgen_2023} & Gaussian primitive  & SDS+MSE & IDB+ESB  & Text/Image  & Consistency\&Controllability  \\ \rowcolor{multi!20}
Free4D~\cite{liu2025free4d} & Gaussian primitive  & L1+LPIPS & ESB   & Text/Image & Consistency\&Controllability\&Fidelity \\ \rowcolor{multi!20}
CoCo4D~\cite{zhou2025coco4d}& Gaussian primitive  & L2+SDS & IDB+ESB  & Text/Image & Consistency\&Controllability\&Fidelity \\ \rowcolor{multi!20}
4Dynamic~\cite{yuan_4dynamic_2024} & Grid  & SDS+MSE & IDB+ESB  & Text/Video  & Consistency\&Diversity  \\ \rowcolor{multi!20}

GenXD~\cite{zhao2024genxd} & NeRF  & SDS & IDB  & Image/Video  & Consistency  \\ \rowcolor{multi!20}
Sync4D~\cite{fu_sync4d_2024} & Gaussian primitive+Material point  & Displacement loss & ESB   & Video+Text/Image  & Consistency\&Controllable\&Fidelity \\ \rowcolor{multi!20}
Human4DiT~\cite{shao_human4dit_2024} & Video  & Sample loss & E2E   & Image+Dynamic SMPL sequences & Consistency\&Controllability\&Efficiency \\ \rowcolor{multi!20}
DreamDrive~\cite{mao2024dreamdrive} & Gaussian primitive & L1+SSIM & ESB & Image+Traj.  & Consistency\&Controllability\\ \rowcolor{multi!20}
Stag-1~\cite{wang2024stag1realistic4ddriving} & Point cloud & MSE  & GDB  & Image+Time+View  & Controllability \\ \rowcolor{multi!20}
OccSora~\cite{wang2024occsora}  & 4D Sequences & KL loss & E2E  & Trajectory Prompt  & Consistency\\\rowcolor{multi!20}
MagicDrive3D~\cite{MagicDrive3D}  & Gaussian primitive & L1+L2 & ESB & Bboxes+BEV map+Traj.+Text & Consistency\&Controllability\\ \rowcolor{multi!20}
\bottomrule


\end{tabular}} \label{tab:all}
\vspace{-20pt}
\end{table*}

To systematically understand the landscape of 4D generation, this section introduces a detailed taxonomy of existing approaches. We categorize methods by conditioning inputs, including text, images, videos, and 3D objects. 
Figure~\ref{fig:control} visually summarizes representative methods and their 4D outputs. 
Within each input category, we further analyze key methods based on their underlying 3D representations: 
{Mesh}~\cite{rossignac1999edgebreaker}, {NeRF}~\cite{mildenhall2021nerf}, {PointCloud}~\cite{levoy1985use}, and {Gaussian Splatting}~\cite{kerbl3Dgaussians}, 
with a comparative summary provided in Table~\ref{tab:all}.


\subsection{Text-to-4D Generation}
Text-to-4D generation leverages textual inputs to synthesize 4D content, {enabling intuitive semantic control and highly flexible user-driven creative workflows.} 
This paradigm uniquely combines natural language semantics with 4D representations, making it ideal for semantic-driven dynamic content creation, text-guided virtual asset generation, and intuitive text-driven character animation/interaction.
Current methods primarily employ diffusion-based generative models and cross-modal optimization frameworks, leveraging diffusion priors from text-to-image, text-to-video, and text-to-3D tasks to guide temporally coherent and semantically accurate 4D generation.

Despite these advancements, constructing sufficiently large-scale and annotated Text-to-4D datasets remains challenging. 
As a result, current approaches typically rely on implicit distillation strategies rather than direct end-to-end training. 
Existing implicit optimization methods~\cite{singer2023text, zheng_unified_2024, li_dreammesh4d_2024, yang2024deformable, wu20244d, yu_4real_2024} leverage multiple viewpoints and timestep diffusion models, indirectly optimizing geometry and motion via SDS.
{Explicitly supervised methods, such as PLA4D~\cite{miao_pla4d_2024}, initialize 4D generation from text-to-3D results and explicitly integrate motion guidance from text-to-video models.}
{
These implicit and explicit optimization strategies have been successfully instantiated within various representation frameworks, including explicit mesh-based, implicit NeRF-based, and differentiable Gaussian-based methods. Each method possesses unique strengths and suitable application scenarios.
}


\textbf{Mesh-based text-to-4D generation methods.} Mesh-based representations provide explicit geometric control, making them particularly suitable for precise text-driven shape manipulation and animation. 
{For instance,} CT4D~\cite{chen_ct4d_2024} implements a three-stage Generate-Refine-Animate pipeline, initially generating a 3D mesh from text prompts, refining textures, and subsequently animating mesh vertices to create dynamic sequences. 
AvatarCLIP~\cite{hong2022avatarclip} introduces a zero-shot method, first generating a static 3D avatar before animating it through registration to an SMPL-based skeletal structure. 

\textbf{NeRF-based text-to-4D generation methods.} NeRF-based approaches implicitly encode continuous geometry and appearance, providing superior flexibility in representing complex temporal variations described by text. 
MAV3D~\cite{singer2023text}, a pioneering NeRF-based approach, employs multi-stage SDS optimization with text-to-image and text-to-video diffusion models, progressively refining textures, geometry, and motion. 
Dream-in-4D~\cite{zheng_unified_2024} proposes a similar two-stage strategy, first employing a multiview diffusion model for static 3D generation, then integrating video diffusion models to enrich temporal dynamics within the NeRF framework.

\textbf{Gaussian Splatting-based text-to-4D generation methods.}
Gaussian splatting techniques support efficient, differentiable rendering, facilitating rapid, high-quality text-driven dynamic scene synthesis. Following the introduction of 4D Gaussian Splatting~\cite{li_dreammesh4d_2024, yang2024deformable, wu20244d}, Ling \textit{et al.} developed AYG~\cite{ling2024align}, a two-stage text-to-4D pipeline. AYG first generates a static 3D Gaussian representation from text using SDS optimization, then enriches dynamic effects through text-to-video diffusion models such as AYL~\cite{blattmann2023align}. Comp4D~\cite{xu_comp4d_2024} further enhances Gaussian representations by integrating large language models (LLMs) to automatically generate multiple dynamic objects and their coherent motion trajectories based on textual inputs, incorporating these into unified scenes.

Recent Gaussian-based methods shift towards explicitly supervised frameworks. PLA4D~\cite{miao_pla4d_2024} introduces a pixel-aligned pipeline, explicitly aligning generated 4D Gaussian representations with video targets, significantly improving control and alignment quality. Similarly, 4Dynamic~\cite{liu2024dynamic} leverages generated videos as explicit reference inputs, optimizing temporal coherence and realism. 4Real~\cite{yu_4real_2024} modifies the Snap Video Model to generate multiview video frames, enabling photorealistic scene-level generation when combined with existing 4D reconstruction techniques. AvatarGO~\cite{cao_avatargo_2024} presents a zero-shot method specifically designed for 4D full-body human-object interactions (HOI), using text-to-3D generated human models animated via the linear blend skinning function of SMPL-X~\cite{loper2023smpl,pavlakos2019expressive}. These approaches highlight explicit motion supervision, advancing Gaussian-based text-to-4D generation capabilities significantly.

\vspace{-5pt}
\subsection{Image-to-4D Generation}
Image-to-4D generation, such as Diffusion²~\cite{yang_diffusionmbox2_2024} and EG4D~\cite{sun_eg4d_2024}, leverages single images to synthesize 4D content, effectively combining spatial information extracted from static inputs with the temporal priors of video models for 4D generation. Despite considerable advancements, the limited availability of dedicated datasets remains a significant challenge, prompting existing methods to adopt two primary approaches. {Methods relying on diffusion model priors }, such as Animate124~\cite{zhao_animate124_2023}, and 4K4DGen~\cite{li_4k4dgen_2024}, utilize priors from image-conditioned diffusion models and indirectly optimize geometry and motion through Score Distillation Sampling (SDS). In contrast, {methods employing explicit supervision from other modalities}, exemplified by DreamGaussian4D (DG4D)~\cite{ren2023dreamgaussian4d}, explicitly use generated video sequences as direct supervisory signals, significantly improving computational efficiency and quality. These explicit methods predominantly employ Gaussian Splatting representations, leveraging efficient differentiable rendering.

\textbf{Gaussian Splatting-based image-to-4D generation methods.} Due to their efficient differentiable rendering capabilities, compact representations, and effectiveness in synthesizing high-quality dynamic scenes from single images, Gaussian Splatting techniques have gained prominence in this direction. Specifically, DreamGaussian4D (DG4D)~\cite{ren2023dreamgaussian4d} pioneers this direction by introducing a two-stage pipeline involving initial static Gaussian splat generation and subsequent dynamic Gaussian deformation. The dynamic stage incorporates temporal enhancement via video-conditioned texture refinement, effectively balancing computational efficiency and realism. Similarly, Diffusion²~\cite{yang_diffusionmbox2_2024} combines priors from multiview and video diffusion models and introduces Variance-Reducing Sampling (VRS) to reconcile differences across heterogeneous diffusion priors, thereby significantly enhancing multiview and temporal consistency. Human4DiT~\cite{shao_human4dit_2024} addresses the challenging scenario of synthesizing dynamic, spatiotemporally coherent human models from single images. It employs a hierarchical 4D transformer architecture that explicitly integrates human-specific features, temporal signals, and camera parameters, providing precise conditioning and robust results. Additionally, EG4D~\cite{sun_eg4d_2024} presents an approach that circumvents the computationally intensive SDS optimization process. EG4D sequentially utilizes image-to-video and image-to-multiview diffusion models to produce temporally and spatially coherent image sequences, subsequently guiding effective dynamic scene reconstruction. 

Further enhancing immersion, 4K4DGen~\cite{li_4k4dgen_2024} specifically focuses on converting panoramic images into dynamic 4D environments. This method effectively maintains spatial and temporal consistency through careful panoramic video diffusion, enabling the efficient generation of immersive and temporally coherent virtual environments. Finally, Disco4D~\cite{pang_disco4d_2024} uniquely targets detailed dynamic human animation by distinctly modeling clothing via Gaussian splatting and human body shapes using the SMPL-X model. By leveraging separate modeling pathways, Disco4D achieves significantly enhanced visual fidelity and improved generation flexibility.

Beyond purely Gaussian-based representations, Animate124~\cite{zhao_animate124_2023} exemplifies a hybrid strategy by initially generating static NeRF-based geometry from single images and subsequently refining temporal dynamics through SDS-optimized video diffusion, providing insights into potential hybrid representation strategies. 
Collectively, these methods have established a robust framework for image-to-4D generation, marking substantial advancements in computational efficiency, realism, and controllability, and suggesting promising avenues for further exploration.

\vspace{-8pt}
\subsection{Video-to-4D Generation}
Video-to-4D generation extends single-view video sequences to multiview videos, synthesizing coherent spatio-temporal content with consistent geometry and motion. 
This task uniquely bridges the gap between video content and dynamic 4D scene representation, significantly advancing applications that require realistic multiview experiences and temporally coherent animations. 
Due to the difficulty in collecting large-scale annotated video-to-4D datasets, current approaches typically adopt { SDS-based} or {multimodal-prior-based} frameworks, often leveraging existing video-conditioned diffusion priors to guide 4D generation.


The former category of methods, such as 4Diffusion~\cite{zhang_4diffusion_2024} and SV4D~\cite{xie_sv4d_2024}, adapt pretrained video diffusion models to 4D generation by fine-tuning them on curated synthetic datasets and leveraging Score Distillation Sampling~\cite{hoppe2022diffusion} to indirectly supervise dynamic geometry and motion coherence. In parallel, the latter class of methods, exemplified by DreamMesh4D~\cite{li_dreammesh4d_2024} and SC4D~\cite{wu_sc4d_2024}, integrate explicit geometric priors and linear blend skinning (LBS) methods to provide robust and direct constraints, significantly improving spatial-temporal coherence and computational efficiency. Recent approaches such as Efficient4D~\cite{liu_efficient_2024} generate multiview data to directly support 4D model training, offering alternative solutions beyond implicit or explicit optimization strategies. Below, we further categorize representative methods according to their representation frameworks: {Mesh-based, NeRF-based, and Gaussian Splatting-based approaches.}

\textbf{Mesh-based video-to-4D generation methods.} Mesh representations explicitly model geometry, making them highly interpretable and amenable to precise animation control. 
DreamMesh4D~\cite{yang_sketch} leverages an initial coarse mesh obtained from an image-to-3D pipeline and constructs a sparse deformation graph to efficiently control mesh deformation over time. 
By integrating linear blend skinning (LBS) with dual-quaternion skinning (DQS), DreamMesh4D achieves notably superior spatial-temporal consistency and computational efficiency compared to implicit volumetric methods such as NeRF or Gaussian Splatting, highlighting the strengths of mesh-based explicit control.

\textbf{NeRF-based video-to-4D generation methods.} 
NeRF representations implicitly encode continuous geometry and appearance, effectively modeling complex volumetric phenomena captured in videos. Consistent4D~\cite{jiang2023consistent4d} first densifies sparse monocular video data through temporal and multiview interpolation, providing explicit supervisory data. Leveraging these enriched datasets, it employs SDS optimization driven by diffusion priors to generate coherent and detailed 4D NeRF representations. Similarly, 4Diffusion~\cite{zhang_4diffusion_2024} introduces a learnable motion module within the diffusion model, specifically tailored for generating consistent multiview videos. After training on synthetic multiview sequences, it converts monocular input into comprehensive multiview outputs, facilitating efficient NeRF-based SDS optimization. SV4D~\cite{xie_sv4d_2024} further enhances this paradigm by constructing diffusion models conditioned explicitly on main-view videos and initial-moment multiview videos, enabling effective completion of missing views and frames, subsequently guiding robust 4D NeRF generation.

\textbf{Gaussian Splatting-based video-to-4D generation methods.} 
Due to their efficient differentiable rendering, compact representations, and effectiveness in rapid dynamic scene synthesis, Gaussian Splatting-based approaches have gained significant attention. Efficient4D~\cite{pan2024efficient4d} notably employs SyncDreamer~\cite{liu2023syncdreamer} to efficiently convert single-view video sequences into temporally coherent multiview image matrices, subsequently enabling rapid and effective 4D reconstruction. Additionally, SC4D~\cite{wu_sc4d_2024} adopts a two-stage pipeline, first initializing sparse control Gaussians to capture shape and motion, and then using these control points to implicitly guide dense Gaussian splats via linear blend skinning, significantly enhancing visual fidelity and motion coherence.

\vspace{-10pt}
\subsection{3D-to-4D Generation}
3D-to-4D generation expands static 3D models into dynamic sequences, synthesizing temporally coherent changes from an initial spatial representation. 
This task uniquely focuses on transforming static geometry into animated sequences, significantly enriching the realism and applicability of existing 3D assets. 
Despite promising potential, research in this direction remains relatively underexplored, primarily emphasizing explicit geometric control via mesh-based methods and implicit neural modeling via NeRF-based approaches.

\textbf{Mesh-based 3D-to-4D generation methods.} Mesh-based representations explicitly encode geometry, facilitating precise control over shape deformation and motion. 
Otberdout \textit{et al.}~\cite{otberdout_3d_2021} propose Motion3DGAN combined with a Sparse2Dense Mesh Decoder, predicting displacements of 3D facial landmarks based on provided expressions. 
Subsequently, these landmark displacements guide mesh deformation, generating expressive and temporally coherent sequences. 
Leveraging a GAN framework, their method efficiently incorporates temporal deformation into static 3D facial structures.

\textbf{NeRF-based 3D-to-4D generation methods.} 
NeRF-based methods implicitly model volumetric geometry, effectively capturing complex temporal variations in dynamic scenes. 
HyperDiffusion~\cite{erkocc2023hyperdiffusion} utilizes a transformer-based diffusion model applied directly to pretrained neural field parameters, enabling the generation of new dynamic NeRF representations. 
Dynamic meshes are subsequently extracted via techniques such as Marching Cubes. Additionally, ElastoGen~\cite{feng2024elastogen} introduces a physically informed method for generating realistic elastodynamic behaviors. 
Initially rasterizing a static 3D model, ElastoGen employs recurrent neural networks~\cite{RNN} and 3D convolutional mechanisms to iteratively predict and refine physically plausible deformation fields, achieving robust temporal coherence.

\vspace{-10pt}
{\subsection{Multi-conditional 4D Generation}}
Multi-conditional 4D generation integrates diverse control modalities—such as text, images, videos, sketches, and physics-based priors—to significantly enhance the controllability and flexibility of dynamic scene synthesis. 
These methods typically combine implicit optimization strategies with explicit supervision, tailored specifically to diverse conditions, enabling fine-grained manipulation across varied user-driven scenarios. 
For instance, Generative Rendering~\cite{cai_generative_2024} employs 4D meshes alongside pretrained diffusion models to produce temporally coherent dynamics, while TC4D~\cite{bahmani_tc4d_2024} introduces precise user control through explicit spline trajectories, guiding both global rigid transformations and local deformations via text-to-video supervision.

In human animation scenarios, Beyond Skeletons~\cite{yang_beyond_2024} combines text and image conditions for enhanced dynamic control, while STAR~\cite{chai_star_2024} employs skeleton-aware Score Distillation Sampling (SDS) and motion retargeting techniques to ensure visual coherence and accuracy.
Physics-driven approaches such as Sync4D~\cite{fu_sync4d_2024} and Phy124~\cite{lin_phy124_2024} leverage differentiable Material Point Method (MPM) simulations to embed realistic motion dynamics derived from video references or textual prompts, significantly boosting physical plausibility and computational efficiency. 
Sketch-to-4D~\cite{yang_sketch} further facilitates intuitive interaction by integrating sketches and textual guidance within a controlled SDS framework enhanced by spatiotemporal consistency modules.
Diffusion4D~\cite{liang_diffusion4d_2024} exemplifies robust multimodal control by adapting text-to-3D and image-to-3D models to dynamic datasets, effectively addressing comprehensive 4D generation scenarios.
By effectively combining multiple conditioning signals, such as text, images, sketches, and physics-based priors, these methods jointly enhance controllability and realism, significantly extending the interactive capabilities of 4D generation.
Future research should deepen multimodal integration, develop adaptive optimization, and enhance computational efficiency for improved controllability and realism in 4D generation.

\ding{42} \hl{Control modalities in 4D generation have progressively evolved from text and images toward video inputs, driven by video’s intrinsic ability to capture rich spatiotemporal dynamics. 
In contrast, the complexity and scarcity of high-quality 3D targets have limited their broader use for direct conditioning. Recent trends increasingly favor hybrid multi-modal control strategies, integrating diverse input sources to enhance fidelity, detail, and precision, thus effectively meeting the demands of specific downstream tasks and applications.}

\begin{figure*}
    \centering
    \includegraphics[width=1\linewidth]{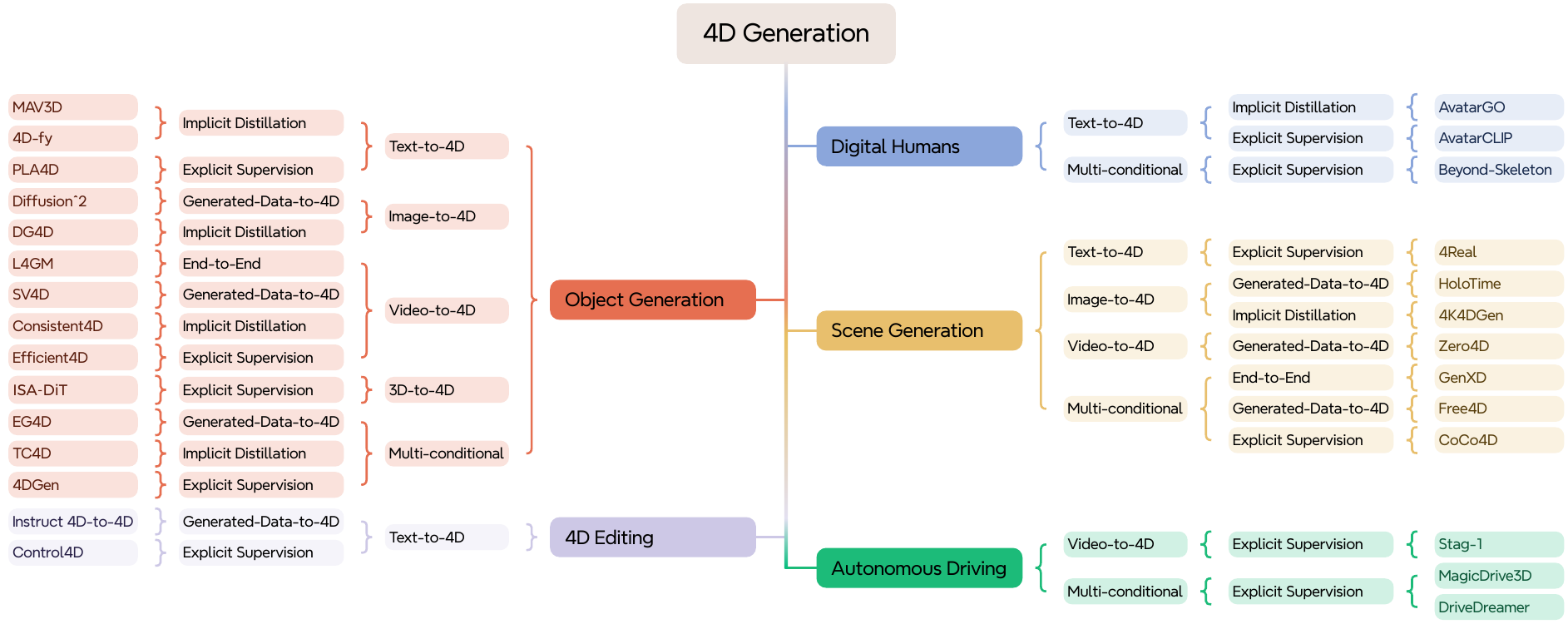}
    \vspace{-20pt}
\caption{\textbf{Illustration of five key applications of 4D generation}: 
(A) \textcolor{text!100}{Object Generation}, synthesizing dynamic objects with detailed geometry and appearance for robotics, simulations, and recognition tasks; 
(B) \textcolor{image!100}{Scene Generation}, reconstructing complex dynamic environments for immersive VR/AR and digital twins; 
(C) \textcolor{video!100}{Digital Humans}, creating realistic human avatars with coherent motion and expressions, suitable for gaming and telepresence; 
(D) \textcolor{3d!100}{4D Editing}, enabling intuitive manipulation of dynamic content, including motion and appearance adjustments over time; and 
(E) \textcolor{multi!100}{Autonomous Driving}, enhancing vehicle perception and decision-making by generating realistic dynamic scenarios. 
Together, these applications showcase the significant potential of 4D generation in bridging spatial and temporal dimensions across diverse practical and research contexts.}
    \label{fig:apps}
    \vspace{-20pt}
\end{figure*}

\vspace{20pt}
\section{Emerging Applications of 4D Generation} \label{sec:4Dappl}
4D generation extends 3D generation by explicitly incorporating the temporal dimension, enabling the synthesis of content that dynamically evolves over time. 
This capability is pivotal for creating realistic virtual environments in VR/AR, delivering advanced visual effects and interactive experiences in film and gaming, and enhancing robotic perception and adaptability within dynamic scenarios. 
As illustrated in Figure~\ref{fig:apps}, current research in 4D generation primarily focuses on five core applications: object generation, scene generation, digital humans, content editing, and autonomous driving.

\vspace{-5pt}
\subsection{4D Object Generation}
The generation of 4D objects refers to synthesizing dynamic, single-object models with coherent spatial and temporal dimensions, effectively integrating motion and geometry under specified conditions. 
Compared to static 3D or conventional video generation, this task uniquely requires maintaining both spatial consistency across viewpoints and temporal continuity throughout dynamic sequences.
Currently, 4D object generation faces challenges in terms of consistency, controllability, fidelity, and efficiency.

4D object generation faces core challenges in achieving temporal and spatial consistency, controllability, fidelity, and computational efficiency. Temporal issues stem from video diffusion models, leading to motion discontinuities and limited output durations, while spatial inconsistencies (the "Janus problem") arise from 2D diffusion models' limited 3D awareness.
While recent methods leverage SDS for spatial coherence and multiview video diffusion for consistency across viewpoints and sequences~\cite{ling2024align,bahmani20244dfy,zheng_unified_2024,wu2024cat4d,xie_sv4d_2024}, scalability and computational costs persist. Beyond consistency, achieving high fidelity and fine-grained control is critical. Current reliance on text, image, or video inputs~\cite{yuan_4dynamic_2024,li_4k4dgen_2024,bahmani_tc4d_2024} presents limitations in spatial precision, static perspectives, or consistent dynamics. Furthermore, prevalent volume-based/neural rendering or overly dense polygonal meshes~\cite{li_dreammesh4d_2024,chen2024meshanything} hinder compatibility with rasterization and computational efficiency.
Despite advancements like DG4D~\cite{ren2023dreamgaussian4d} significantly accelerating generation, challenges remain in scalability, real-time deployment, multimodal semantic control, and robust evaluation, necessitating continued exploration of efficient representations and hybrid frameworks.

\vspace{-5pt}
\subsection{4D Scene Generation}
4D scene generation involves synthesizing dynamic scenes composed of multiple interacting targets within a coherent spatiotemporal environment. Unlike 4D object generation, which prioritizes localized details of individual objects, scene generation emphasizes the overall dynamic interplay among multiple objects and their interaction with complex backgrounds.

Several approaches have addressed the challenges inherent in 4D scene generation, particularly regarding spatial and temporal coherence. For instance, 4K4DGen~\cite{li_4k4dgen_2024} enables the creation of large-scale panoramic dynamic scenes, allowing flexible user navigation and viewpoint adjustment. 
Similarly, Generative Rendering~\cite{cai_generative_2024} synthesizes datasets containing multiple interacting targets and detailed backgrounds, which can directly guide subsequent 4D reconstruction methods, ensuring strong spatiotemporal consistency. Moreover, 4Real~\cite{yu_4real_2024} utilizes intermediate video outputs (reference and freeze-time videos) derived from text inputs to effectively guide scene-level 4D generation, enhancing visual realism and immersion.\footnote{Certain methods claimed as 4D scene generation approaches~\cite{yang_sketch,xu_comp4d_2024,singer2023text} predominantly focus on isolated objects with minimal environmental interaction and simplistic backgrounds, making them more suitable for object-centric tasks rather than comprehensive scene generation tasks.}
These recent advances highlight substantial progress in 4D scene generation, yet further efforts remain necessary to robustly synthesize complex and immersive dynamic scenes featuring detailed inter-object interactions and fully realized environmental contexts.

\vspace{-10pt}
\subsection{4D Digital Human Generation}
4D digital human generation involves synthesizing highly detailed, temporally evolving, and interactive virtual human avatars from various input modalities such as text, images, and skeleton data. 
Unlike traditional dynamic human reconstruction, which typically relies on extensive multiview video data~\cite{zhao2022human, icsik2023humanrf}, 4D digital human generation methods focus on generative models and data-driven frameworks, enabling flexible and customizable synthesis of realistic human avatars with coherent spatial and temporal dynamics.

Mesh-based methods, exemplified by AvatarCLIP~\cite{hong2022avatarclip}, leverage explicit geometric representations to generate detailed and animatable 3D avatars driven by textual descriptions. 
AvatarCLIP specifically enables customized avatar creation and natural-language-driven animations, achieving realistic textures and vivid motions. 
Gaussian-based approaches, such as GAvatar~\cite{yuan2024gavatar}, employ neural implicit fields to predict Gaussian attributes efficiently, significantly enhancing animation stability and learning efficiency.

Additionally, image-driven methods have also emerged, exemplified by Human4DiT~\cite{shao_human4dit_2024}, which utilizes hierarchical 4D transformer architectures to integrate spatiotemporal attention mechanisms. 
This method generates high-quality, coherent 360-degree human animations from single-view images by effectively modeling spatial and temporal dynamics. 
Further advancing the field, AvatarGO~\cite{cao_avatargo_2024} specifically targets realistic human-object interactions (HOI), leveraging large language model guidance and correspondence-aware optimization to produce animatable 4D HOI scenes with robust realism.
Skeleton-driven methods, such as STAR~\cite{osman2020star}, have also significantly contributed by incorporating skeleton-aware Score Distillation Sampling (SDS) optimization and motion retargeting. 
These techniques effectively address common alignment and coherence issues, refining surface details and enhancing the spatial-temporal fidelity of generated avatars.

Collectively, these advancements in 4D digital human generation have significantly elevated realism, stability, and controllability, transforming the landscape of interactive virtual human synthesis. 
Nonetheless, key challenges remain, particularly in achieving real-time computational efficiency, diversifying realistic motions, seamlessly integrating multimodal inputs, and enhancing user-driven interactivity.

\vspace{-10pt}
\subsection{4D Content Editing}
4D content editing involves generating or modifying dynamic 3D content within specified regions guided by user-provided instructions or prompts. 
This editing can target local or global regions, encompassing tasks such as object addition and removal, attribute modification, and style transfer, each requiring different degrees of spatial and temporal coherence.

Recent methods have significantly advanced 4D editing capabilities. 
Instruct 4D-to-4D~\cite{mou2024instruct}, a pioneering pseudo-3D editing method, utilizes Instructpix2pix~\cite{brooks2023instructpix2pix} for anchor image modification and employs optical flow to propagate these edits throughout the scene. 
While effective for stylization, it struggles with object addition, removal, and complex attribute changes due to its dependence on 2D image editing techniques. 
Conversely, 4D-Editor~\cite{jiang20234d} introduces hybrid radiance and semantic fields distilled from the DINO~\cite{caron2021emerging} teacher model, allowing interactive editing via user-marked reference views. 
This method excels at precise object-level edits but is less effective for fine-grained attribute adjustments and stylization tasks. 
Control4D~\cite{shao2024control4d} addresses human portrait editing through GaussianPlanes integrated with GAN-based generators and 2D diffusion editors, enabling consistent, high-quality temporal modifications based purely on textual instructions.

Despite these advancements, significant challenges remain, particularly regarding the maintenance of spatiotemporal consistency. 
Ensuring temporal continuity across frames and spatial coherence across viewpoints poses complex requirements. 
Addressing these dual demands calls for sophisticated algorithms capable of seamlessly integrating dynamic modifications while preserving overall coherence. 
Future research may benefit from focusing on improved algorithmic robustness, scalability, and interactivity, thereby supporting broader applications such as immersive media, virtual reality, and interactive digital environments.

\vspace{-10pt}
\subsection{4D Generation for Autonomous Driving}
Autonomous driving demands precise perception and planning in dynamic environments, yet synthesizing realistic and temporally coherent visual scenes remains challenging. 
Traditional methods often struggle to produce high-fidelity dynamic scenarios, motivating the adoption of 4D generation techniques. 
By integrating spatial and temporal modeling, these techniques substantially enhance the simulation and prediction capabilities of autonomous systems.
For example, MagicDrive3D~\cite{MagicDrive3D} uses diffusion priors for dynamic scene generation, while DreamDrive~\cite{mao2024dreamdrive} leverages video diffusion models and hybrid Gaussian representations to boost consistency and quality. Similarly, Stag-1~\cite{wang2024stag1realistic4ddriving} integrates sparse point cloud reconstructions with video diffusion to enhance viewpoint diversity and temporal continuity.

Furthermore, the development of world models~\cite{ha2018world} has expanded autonomous vehicle capabilities for sophisticated perception and planning, heavily relying on advanced 4D generation for synthesizing realistic driving scenarios to facilitate predictive modeling and strategic decision-making. DriveDreamer~\cite{wang2024drivedreamer}, for instance, constructs world models directly from real-world data for accurate traffic dynamics, and OccSora~\cite{wang2024occsora} introduces diffusion-based methods for comprehensive dynamic 4D occupancy representations, significantly improving temporal stability in longer scenes. Despite these advances, significant challenges persist in 4D generation for autonomous driving, notably achieving consistent, high-fidelity novel-view synthesis and maintaining temporal coherence over extended sequences. Addressing these barriers is crucial for reliable and scalable deployment; thus, future research must prioritize enhancing generative fidelity, extending temporal modeling capabilities, and improving real-time computational efficiency.

\ding{42} \hl{ Recent advancements in 4D generation have substantially enhanced capabilities across diverse domains, including single-object modeling, comprehensive scene synthesis, interactive digital humans, controllable content editing, and autonomous driving simulations. 
Despite notable progress in realism, consistency, and interactivity, common challenges persist—particularly in computational efficiency, scalability for complex scenes, multimodal integration, and robust spatiotemporal coherence. 
Addressing these shared limitations will be pivotal for translating the promising potential of 4D generation into practical, widely deployable technologies.}

\begin{figure*}[ht]
    \centering
    \includegraphics[width=1\linewidth]{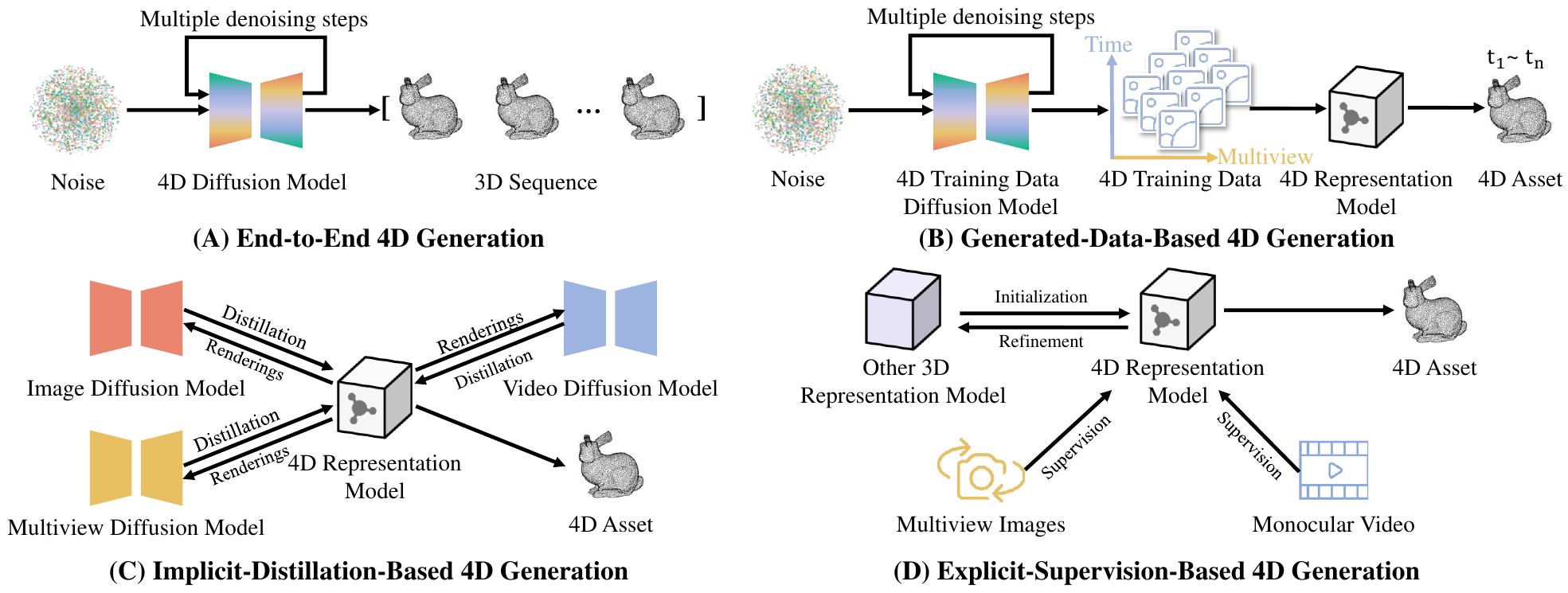}
    \vspace{-20pt}
    \caption{\textbf{A summarization of four paradigms commonly used in current 4D generation methods. }
(A) {End-to-End 4D Generation}: Directly synthesizing 4D assets via a trained 4D diffusion model. 
(B) {Generated-Data-Based 4D Generation}: Generating multiview, multi-temporal training data with diffusion models to indirectly facilitate 4D reconstruction. 
(C) {Implicit-Distillation-Based 4D Generation}: Employing multiple diffusion models through implicit distillation, providing generative priors for 4D assets. 
(D) {Explicit-Supervision-Based 4D Generation}: Utilizing explicitly generated multimodal data to provide direct supervisory signals for constructing 4D representations.}
    \label{fig:base}
    \vspace{-20pt}
\end{figure*}

\section{Paradigm of 4D Generation} \label{sec:paradigms}
While current 4D generation methods exhibit considerable methodological overlap, this section systematically categorizes existing approaches into four principal paradigms: End-to-End 4D generation (E2E), Generated-Data-Based 4D generation (GDB), Implicit-Distillation-Based 4D generation (IDB), and Explicit-Supervision-Based 4D generation (ESB), as illustrated in Figure~\ref{fig:base}. 
Table~\ref{tab:all} further summarizes and annotates existing 4D generation methods according to these identified paradigms, clarifying their methodological distinctions and commonalities.

\vspace{-12pt}
\subsection{End-to-End 4D Generation}
End-to-end 4D generation refers to the direct synthesis of 4D assets from inputs using a single model, without intermediate manual intervention or complex inter-module connections. 
{Consequently, some current methods~\cite{ren2024l4gm,deng2025stp4d, wang2024occsora} have achieved  large-scale, end-to-end 4D generation models.}
NVIDIA's L4GM~\cite{ren2024l4gm} stands as the first network supporting end-to-end 4D generation. 
It employs a Transformer-based architecture specifically designed with temporal self-attention to capture temporal variations and cross-view self-attention to capture view changes. 
The network is trained using a reconstruction loss on the curated Objaverse-4D dataset. 
During inference, it generates a 3D Gaussian representation for each video frame, with intermediate time steps filled by interpolation.

\vspace{-12pt}
\subsection{Generated-Data-Based 4D Generation}
Generated-data-based 4D generation refers to the approach where multiview, multi-temporal images are first synthesized to form a training dataset. These generated images are then directly used with existing 4D representation models for reconstruction.
A prominent example of this paradigm is SV4D~\cite{xie_sv4d_2024}. This method extends SVD (a video generation model) by incorporating spatial attention, view attention, and frame attention modules to model spatiotemporal variations. 
SV4D is trained on the ObjaverseDy dataset, and during inference, it employs a mixed-sampling strategy to enhance spatiotemporal consistency among the generated data. 
Subsequently, the authors use these generated images to reconstruct the final 4D targets, leveraging a NeRF-based representation.

\vspace{-10pt}
\subsection{Implicit-Distillation-Based 4D Generation}
Implicit-Distillation-Based 4D generation primarily leverages the Score Distillation Sampling (SDS) algorithm, or its enhanced variants, to generate 4D assets through a distillation process.
MAV3D~\cite{singer2023text} stands as a prime example of this paradigm. 
It pioneered the application of DreamFusion's SDS to 4D generation, utilizing image diffusion models, video diffusion models, and super-resolution video diffusion models to provide priors for generating 4D targets.
{These methods are characterized by an iterative optimization loop, in which renderings from the evolving 4D representation are repeatedly fed into multiple diffusion models. 
This process indirectly supervises the 4D model’s geometry and motion, effectively injecting diffusion-based priors through rendered feedback.}


\vspace{-10pt}
\subsection{Explicit-Supervision-Based 4D Generation}
Explicit-Supervision-Based 4D Generation primarily relies on explicit supervision provided by inputs such as images, videos, or 3D data. 
This includes, but isn't limited to, methods that use image supervision or 3D object initialization for 4D generation.
EG4D~\cite{sun_eg4d_2024} is a leading example of this paradigm. 
It introduces an attention injection strategy for synthesizing temporally consistent multiview videos, a robust and efficient dynamic reconstruction method based on Gaussian Splatting, and a refinement stage with a diffusion prior for semantic restoration, all to generate 4D targets purely through explicit supervision.

\begin{table}[tp]
    \centering
    \caption{\textbf{4D generation paradigms comparison.}  We highlight the key differences among End-to-End (E2E), Generated-Data-Based (GDB), Implicit-Distillation-Based (IDB), and Explicit-Supervision-Based (ESB) 4D generation. }
    \vspace{-5pt}
\resizebox{1\linewidth}{!}{
    \begin{tabular}{c|ccccc}
    \toprule
    \cellcolor{white} 
    
    Paradigms &   Consistency&Controllability&Diversity&Efficiency&Fidelity \\
    \midrule
    E2E~\cite{ren2024l4gm}  & \ding{73} & \ding{73}  & \ding{73}  & \ding{73}\ding{73}\ding{73}  & \ding{73}  \\
    GDB~\cite{xie_sv4d_2024,park2025zero4d}  & \ding{73}\ding{73}  & \ding{73}  & \ding{73}  & \ding{73}\ding{73}   & \ding{73}\ding{73}  \\
    IDB~\cite{singer2023text,yin_4dgen_2023}  & \ding{73}\ding{73} & \ding{73}\ding{73}\ding{73}   & \ding{73}\ding{73}\ding{73} & \ding{73}  & \ding{73}\ding{73}  \\
    ESB~\cite{sun_eg4d_2024,pan2024efficient4d} & \ding{73}\ding{73}\ding{73} & \ding{73}\ding{73} & \ding{73}\ding{73}  & \ding{73}\ding{73}  & \ding{73}\ding{73}\ding{73}  \\
    
    \bottomrule
    \end{tabular}}
    \label{tab:diff}
    \vspace{-15pt}
\end{table}

\ding{42} \hl{ Current 4D generation methodologies broadly follow four primary paradigms, spanning from direct end-to-end generation and intermediate data-driven approaches to implicit distillation and explicit supervision strategies, as shown in Figure}~\ref{fig:base}.
\hl{Table}~\ref{tab:diff} \hl{further provides a comparative analysis of the differences and advantages among these paradigms.}
\hl{End-to-End (E2E) methods offer high computational efficiency, though there remains considerable potential to enhance generation quality. 
Generated-Data-Based (GDB) methods effectively balance efficiency and quality, although controllability could be further improved. 
Implicit-Distillation-Based (IDB) methods, leveraging the strong generative power of diffusion models, currently achieve the highest generation quality and offer extensive control conditions, but at the cost of higher computational demands. 
Explicit-Supervision-Based (ESB) methods provide an effective balance between efficiency and quality, yet expanding the range and flexibility of control inputs remains an area for continued exploration.}
\hl{In practice, researchers increasingly integrate multiple paradigms to exploit complementary advantages and mitigate inherent limitations. }

\section{4D Generation Challenges} \label{sec:4Dcha}
Recent research has enabled significant progress in various 4D generation tasks, supporting the creation of diverse 4D assets ranging from individual objects to complex scenes under multiple control conditions. 
Nonetheless, existing methods predominantly aim at addressing five core technical challenges, summarized comprehensively in Table~\ref{tab:all}: consistency, controllability, diversity, efficiency, and fidelity.

\vspace{-10pt}
\subsection{Geometric and Temporal Consistency in 4D Generation}
Consistency is a critical challenge in 4D generation, encompassing both geometric and temporal aspects. 
Geometric consistency ensures coherent object structures across multiple viewpoints at a single time instance, while temporal consistency maintains smooth transitions across consecutive frames from a fixed viewpoint. 
Effective 4D generation thus requires simultaneous preservation of these two forms of consistency to ensure visual coherence when viewpoints and timestamps vary concurrently.

To enhance consistency, physics-based simulations have been integrated into recent generative frameworks. 
Meng \textit{et al.}~\cite{meng2025grounding} bind static 3D keypoints to physics-driven particle simulations, grounding dynamic transitions in physical principles such as mass and momentum conservation. 
This significantly improves temporal and geometric coherence by ensuring physically plausible motions. 
Complementarily, direct optimization of neural architectures and diffusion models also addresses consistency challenges. 
Methods such as 4Diffusion~\cite{zhang_4diffusion_2024} fine-tune existing diffusion frameworks (\textit{e.g.}, ImageDream~\cite{wang2023imagedream}) explicitly for temporal consistency, while 4K4DGen~\cite{li_4k4dgen_2024} incorporates depth priors into diffusion processes to enhance geometric coherence across viewpoints. 
Additionally, Generative Rendering introduces improved self-attention mechanisms and correspondence-aware feature blending for better cross-frame consistency. 
Similarly, L4GM~\cite{ren2024l4gm} integrates cross-view self-attention layers within its U-Net architecture to explicitly learn consistency patterns directly from data.

Generating intermediate multiview, multi-temporal image grids as auxiliary datasets for subsequent reconstruction represents another effective strategy. 
Diffusion²~\cite{yang_diffusionmbox2_2024} employs the Variable Reconciliation Strategy (VRS) to harmonize heterogeneous scores during the diffusion process, thereby mitigating inconsistencies across viewpoints and time steps. 
Approaches such as Diffusion4D~\cite{liang_diffusion4d_2024}, Human4DiT~\cite{shao_human4dit_2024}, SV4D~\cite{xie_sv4d_2024}, and 4DGen~\cite{yin_4dgen_2023} further retrain diffusion models on curated 4D datasets, explicitly learning consistency from data-driven priors. 
Additionally, hybrid methods exemplified by Consistent4D~\cite{jiang2023consistent4d} and 4Real~\cite{yu_4real_2024} combine models such as SVD~\cite{blattmann2023stable} and SV3D~\cite{voleti2024sv3d} with advanced interpolation techniques to achieve enhanced smoothness and coherence.

Integration of complementary 4D representations further contributes to consistency improvements. 
CT4D~\cite{chen_ct4d_2024} and DreamGaussian4D, for instance, combine Gaussian representations—known for smooth geometric transitions—with mesh-based models that offer detailed structural fidelity. 
These integrated approaches demonstrate superior spatial-temporal coherence compared to single-representation methods. Despite these advancements, ensuring robust and scalable consistency remains challenging, particularly across extensive temporal durations and diverse viewpoints. 
Continued research efforts are needed to improve multi-representation integration, computational scalability, and consistency-aware optimization techniques, thereby enabling broader practical applicability and richer interactive experiences in dynamic 4D generation.

\vspace{-10pt}
\subsection{Flexible and Precise Control Strategies}
Controllability in 4D generation refers to the capability of precisely guiding and manipulating both the generation process and the final dynamic outputs. 
Current approaches primarily enhance controllability through two complementary strategies: employing multiple representations and integrating diverse external conditions.

Multi-representation approaches exploit the complementary strengths of different geometric or physical representations to achieve enhanced controllability. 
For instance, DreamMesh4D~\cite{li_dreammesh4d_2024} combines Gaussian and mesh representations by binding Gaussian points to mesh surface patches, enabling Gaussian models to efficiently encode motion dynamics while leveraging mesh structures for visually high-quality rendering. 
Similarly, Sync4D~\cite{fu_sync4d_2024} and Phy124~\cite{lin_phy124_2024} integrate Gaussian representations with physics-based Material Point Method (MPM) simulations~\cite{de2020material,sulsky1994particle}. 
These methods ensure that the generated 4D content not only exhibits a realistic visual appearance but also adheres strictly to physically plausible motion constraints, significantly enhancing user control over physical realism and dynamic plausibility.

In parallel, methods that integrate diverse external conditions provide explicit control through input signals such as textual descriptions, trajectories, sketches, or video references. 
Human4DiT~\cite{shao_human4dit_2024} incorporates human identity, temporal signals, and camera parameters within its hierarchical transformer network, enabling precise and targeted manipulation of generated 4D human videos. 
Generative Rendering~\cite{cai_generative_2024} uses dynamic meshes as explicit input to guide pre-trained text-to-image models, offering precise control over detailed visual aspects. 
For targeted motion control, TC4D~\cite{bahmani_tc4d_2024} conditions generated dynamics on user-defined trajectories, whereas Beyond Skeletons~\cite{yang_beyond_2024} integrates human skeleton structures to accurately refine motion dynamics. 
Additionally, DG4D~\cite{ren2023dreamgaussian4d} and PLA4D~\cite{miao_pla4d_2024} utilize explicit video supervision, effectively guiding both temporal coherence and spatial fidelity. 
Appearance-focused control methods, exemplified by Sketch-2-4D~\cite{yang_sketch} and 4DGen~\cite{yin_4dgen_2023}, extend controllability by generating outputs guided explicitly by user-provided sketches or textual and visual inputs.

Collectively, these strategies significantly enhance the flexibility and precision of 4D generation methods, providing users with more intuitive and fine-grained control capabilities.
Nevertheless, several critical challenges remain, particularly in balancing control granularity with computational efficiency, ensuring robustness under diverse input conditions, and designing user interfaces that seamlessly integrate multi-modal control signals. 

\vspace{-5pt}
\subsection{Variability and Diversity of Generated Content}
Diversity in 4D generation refers to the capability of producing varied yet coherent outputs in attributes such as motion, appearance, shape, and textures. 
Enhancing such diversity significantly influences the versatility and adaptability of generative models, facilitating their deployment in complex generative scenarios. 
Current research primarily addresses diversity through three distinct strategies: joint attribute optimization, diverse conditional inputs, and diversity transfer.

Joint attribute optimization methods simultaneously optimize geometry, texture, and motion attributes within a unified framework, exemplified by STAR~\cite{osman2020star}, which jointly updates geometric structures, textures, and motion dynamics through text-driven guidance, enabling richer output variations compared to conventional two-stage generation approaches. 
In parallel, diverse conditional input methods, such as DG4D~\cite{ren2023dreamgaussian4d} and PLA4D~\cite{miao_pla4d_2024}, transfer diversity from varied intermediate control conditions—especially diverse video inputs—directly into generated 4D content, thus significantly broadening the range of possible motion and visual appearances. 
Furthermore, diffusion model-based diversity transfer approaches, represented by 4Dynamic~\cite{liu2024dynamic}, leverage Score Distillation Sampling (SDS) with pre-trained 2D diffusion models, effectively translating the inherent variability from the 2D image domain into coherent and diverse 4D outputs, thereby enriching shape, motion dynamics, and texture variations.

Despite notable advancements, existing evaluation methodologies still lack robust metrics tailored specifically to measure multi-dimensional diversity. 
Future research should prioritize developing dedicated benchmarks and metrics capable of quantitatively capturing diversity nuances, explore deeper integration of multimodal conditioning, such as combining text-, image-, and sketch-based inputs, and investigate adaptive representation strategies that balance diversity and coherence. 

\vspace{-10pt}
\subsection{Computational Efficiency and Scalability}
Efficiency remains a critical bottleneck in 4D generation, largely due to the prevalent reliance on computationally intensive SDS optimization. 
Recent research has addressed this challenge by synthesizing intermediate training data, adopting explicit yet efficient representations, and integrating intermediate supervision signals.

Generating intermediate datasets directly has been effective in circumventing extensive optimization. 
For instance, Efficient4D~\cite{pan2024efficient4d} synthesizes dense multiview video data for rapid 4D reconstruction. 
Similarly, Diffusion$^{2}$~\cite{yang_diffusionmbox2_2024} constructs comprehensive multi-frame and multiview image matrices from single images, facilitating efficient subsequent generation. 
Diffusion4D~\cite{liang_diffusion4d_2024} further generalizes this strategy by utilizing unified diffusion models trained on curated datasets, directly generating 4D outputs conditioned on text or images.

Explicit representations and intermediate supervision have also proven beneficial in accelerating 4D generation. 
Phy124~\cite{lin_phy124_2024} and Trans4D~\cite{zeng_trans4d_2024} utilize explicit Gaussian-based representations to significantly reduce computational demands and optimize rendering. 
Additionally, 4Real~\cite{yu_4real_2024} introduces selective Gaussian filtering, effectively balancing computational efficiency and visual quality. 
PLA4D~\cite{miao_pla4d_2024} leverages pixel-level guidance from intermediate images or videos, minimizing dependence on iterative SDS optimization and enhancing efficiency without compromising output quality.

Collectively, these efficiency-focused strategies significantly reduce computational demands in 4D generation, enhancing scalability for large-scale and dynamic scenarios.
Nevertheless, effectively balancing efficiency, visual quality, and real-time interactivity remains challenging, warranting future research into adaptive methods and optimized representations.

\vspace{-10pt}
\subsection{Generation Fidelity and Realism}
Fidelity in 4D generation pertains to accurately capturing detailed geometry, motion, and textures in generated content, while closely adhering to specified input conditions and maintaining temporal consistency. 
Improving fidelity is essential to producing realistic, high-quality 4D outputs suitable for practical applications.
Current approaches typically address fidelity through enhancing spatial-temporal coherence and incorporating precise conditional constraints, such as detailed semantic alignment and structural consistency.

To enhance multiview spatial-temporal correlation and achieve accurate input alignment, 4Diffusion~\cite{zhang_4diffusion_2024} leverages the generalization capabilities of generative models to produce coherent multiview video data. SC4D~\cite{wu_sc4d_2024} incorporates explicit constraints via mean squared error (MSE) loss between generated and real videos, ensuring the generated shapes and motions precisely match specified characteristics.

To further strengthen fidelity through detailed input controls, Animate124~\cite{zhao_animate124_2023} incorporates ControlNet-based constraints, rigorously preserving input image conditions. In contrast, Disco4D~\cite{pang_disco4d_2024} focuses specifically on appearance fidelity, separately optimizing Gaussian representations of clothing to maintain consistent body structures while accurately reflecting attire changes. 
Text-guided methods, such as Dream-in-4D~\cite{zheng_unified_2024} and 4D-fy~\cite{bahmani20244dfy}, emphasize semantic and stylistic fidelity by integrating multiple diffusion models, thereby ensuring accurate translation of textual descriptions into visual content. 
Meanwhile, Comp4D~\cite{xu_comp4d_2024} ensures fidelity in complex multi-object scenes by independently optimizing individual object fidelity before combining them into cohesive, high-quality dynamic environments.

These targeted fidelity enhancement strategies collectively advance realism and input adherence in generated 4D content. 
Future research should continue exploring adaptive loss formulations (\textit{e.g.}, dynamically weighted loss functions for selective fidelity enhancement), improved multimodal conditioning strategies (\textit{e.g.}, joint text-video embedding alignment), and robust fidelity preservation techniques to further elevate realism and expand the applicability of 4D generation methods in advanced practical scenarios.

\ding{42} \hl{  Current advancements in 4D generation have primarily addressed core challenges including spatial-temporal consistency, controllability, diversity, computational efficiency, and fidelity. 
Despite significant progress in integrating diverse representations, multimodal inputs, and efficient optimization methods, substantial issues remain, particularly in balancing computational cost and visual quality, improving precise control in complex scenarios, and developing standardized evaluation metrics. 
Addressing these critical challenges will drive broader applicability, enabling practical and immersive applications of 4D generation.}

\section{Comprehensive Discussion and Future Directions} \label{sec:discussion}
The field of 4D generation has achieved remarkable progress in recent years, significantly enhancing capabilities for modeling dynamic spatio-temporal phenomena and broadening its applicability across diverse domains. 
However, several fundamental challenges remain unresolved. Building upon these previously discussed challenges, this section further identifies and discusses critical future directions that require deeper investigation. 
Specifically, we highlight three essential aspects: (1) constructing large-scale and multimodal 4D datasets, (2) developing unified and efficient frameworks for 4D generation, and (3) establishing comprehensive benchmarks tailored specifically to 4D generative tasks.

\subsection{Building Large-scale and Multimodal 4D Datasets}
High-quality datasets have been foundational to recent advancements in generative modeling, notably in the 2D and 3D domains. 
Extending this progress to 4D generation critically depends on the availability of comprehensive and high-quality 4D datasets, which capture both multiview and dynamic temporal information. 
However, current efforts face significant challenges due to the limited availability of comprehensive and well-annotated 4D datasets.

Recent approaches have primarily addressed this issue by adapting and filtering subsets from existing large-scale static 3D repositories, such as Objaverse~\cite{deitke2023objaverse,deitke2024objaverse}, to create dynamic datasets with disentangled viewpoint and temporal variations. 
Table~\ref{tab:data} provides an overview of the datasets currently employed for 4D generation. Diffusion4D~\cite{liang_diffusion4d_2024} curates a dataset comprising 54,000 dynamic assets through systematic filtering based on structural similarity and motion dynamics. 
Similarly, SV4D~\cite{xie_sv4d_2024} introduces ObjaverseDy, employing adaptive sampling strategies and viewpoint adjustments to enhance the quality and usability of dynamic 4D data. 
Consistent4D~\cite{jiang2023consistent4d} built a monocular video dataset, comprising both synthetic and real data, for video-to-4D generation tasks.

Despite these advancements, significant limitations persist, including the relatively small dataset scale, limited modality diversity, and the absence of comprehensive multimodal annotations.
The scale of current 4D datasets remains modest compared to their static counterparts, restricting the generalizability and robustness of 4D generative models. 
Furthermore, existing datasets predominantly consist of synthetic, rendered images without accompanying textual descriptions or other multimodal annotations. 
This significantly limits the advancement of multimodal 4D generative models, restricting their capability to achieve complex interactions guided by natural language or diverse conditional inputs.

Consequently, future research should prioritize constructing larger-scale, multimodal 4D datasets that include detailed textual descriptions and richer semantic annotations. 
These enhanced datasets would enable training more versatile and generalizable 4D models, facilitating research into multimodal generation tasks such as text-driven dynamic content synthesis and interactive 4D scene understanding. 
Such efforts are essential to realizing the full potential of 4D generation technology, supporting diverse and innovative applications across entertainment, virtual reality, and interactive media.

\begin{table}[tp]
    \centering
    \caption{\textbf{Datasets for 4D generation}. Existing 4D datasets are largely characterized by a lack of diversity, often providing only object-centric training data or restricting evaluation to specific conditions. This scarcity underscores the nascent and open nature of the 4D generation task. }
    \vspace{-6pt}
\resizebox{1\linewidth}{!}{
    \begin{tabular}{c|ccccc}
    \toprule
    \cellcolor{white} 
    Dataset & Type  &Year & Condition & Training set & Test set  \\  
    \midrule
    Consistent4D~\cite{jiang2023consistent4d} & Monocular video & 2023 & Video & \ding{51} & \ding{51}  \\  
    Objavarse-XL~\cite{liang_diffusion4d_2024} & 3D Objects & 2024 & Image/Video & \ding{51} & \ding{55}  \\ 
    Objavarse-Dy~\cite{deitke2024objaverse} & 3D Objects & 2024 & Image/Video & \ding{51}  & \ding{55} \\ 
    \bottomrule
    \end{tabular}}
    \label{tab:data}
    \vspace{-18pt}
\end{table}

\vspace{-10pt}
\subsection{Unified Frameworks for Efficient 4D Generation}
Current 4D generation approaches predominantly rely on multistage pipelines, integrating multiple separately trained diffusion models, including image, multiview, and video diffusion models, with score distillation sampling (SDS)~\cite{bahmani20244dfy,miao_pla4d_2024,ling2024align,yin_4dgen_2023,zheng_unified_2024}.
While effective, this fragmented approach introduces notable complexities, substantial computational overhead, and difficulties in achieving strong generalization, limiting practical applicability.

A fundamental limitation of these multi-stage pipelines is the inherent disconnect between spatial and temporal modeling.
The prevalent practice of separately leveraging multiview diffusion models for spatial consistency and video diffusion models for temporal coherence frequently leads to inconsistencies when synthesizing unified 4D assets~\cite{blattmann2023align,chen2023videocrafter1,singer2022make}.
To effectively overcome this issue, future research should focus on developing unified diffusion models capable of simultaneously encoding both spatial (multiview) and temporal priors within a single cohesive framework.
Such models inherently promote cross-view and temporal consistency, thereby improving the quality and coherence of the generated outputs.


Furthermore, current SDS-based optimization methods, despite their versatility, remain computationally expensive, limiting scalability~\cite{shi2023mvdream,liu2023zero1to3}.
Future research should focus on developing more efficient optimization techniques, such as lightweight distillation, distillation-free optimization, or adaptive algorithms, to dynamically balance efficiency and generative quality.
Meanwhile, enhancing the generalization capabilities of models through advanced multimodal conditioning, physics-informed priors, or adaptive regularization will further improve fidelity and robustness across diverse scenarios, reducing sensitivity to training data variations.

Addressing these interconnected challenges—developing unified diffusion models that simultaneously capture spatial and temporal coherence, advancing efficient optimization techniques to reduce computational overhead, and enhancing generalization capabilities through advanced multimodal conditioning—will be crucial for advancing the practicality and scalability of 4D generation. 
These improvements will ultimately facilitate the broader integration of 4D technologies into interactive and immersive applications, significantly enhancing user experiences across various domains, such as virtual reality, digital entertainment, and autonomous systems.

\begin{table}[tp]
    \centering
    \caption{\textbf{Metrics for 4D generation}. Current evaluation metrics are primarily adapted from image and video generation, supplemented by specific task-oriented metrics. }
    \vspace{-6pt}
\resizebox{1\linewidth}{!}{
    \begin{tabular}{c|ccccccc}
    \toprule
    \cellcolor{white} 

    Metric & Text-to-4D  & Image-to-4D  & Video-to-4D  & 3D-to-4D  & Multi-conditional 4D  \\
    \midrule
    PSNR~\cite{PSNR} &\ding{55} &\ding{51} &\ding{51}  &\ding{55} & \ding{55} \\
    LPIPS~\cite{LPIPS} &\ding{55}  &\ding{51} &\ding{51} &\ding{55} &\ding{55} \\
    SSIM~\cite{wang2004image} &\ding{55}   &\ding{51} &\ding{51} &\ding{55} &\ding{55}\\
    FVD~\cite{FVD} &\ding{55} &\ding{51} &\ding{51} &\ding{55}  &\ding{55}\\
    CLIP~\cite{CLIP} &\ding{51} &\ding{51}  &\ding{51} &\ding{55} &\ding{55}  \\
    User Study &\ding{51} &\ding{51} &\ding{51}  &\ding{51} &\ding{51} \\
    Others  &\ding{51}  &\ding{51} &\ding{51} &\ding{51}&\ding{51}\\
    \bottomrule
    \end{tabular}}
    \label{tab:Metric}
    \vspace{-18pt}
\end{table}

\vspace{-10pt}
\subsection{Towards Comprehensive Benchmarks for 4D Generation}
Despite the rapid advancements and increasing diversity of 4D generative models, evaluation practices remain in the early stages of development, with a notable reliance on subjective assessments such as user studies.
While human evaluations provide valuable insights into perceptual quality and user experience, their inherent subjectivity, limited scalability, and inconsistency across different evaluators pose significant obstacles to robust and reproducible model comparisons.

To partially address these limitations, researchers currently utilize a range of objective metrics adapted primarily from image and video generation domains (Table \ref{tab:Metric}). 
Metrics such as PSNR~\cite{PSNR} and SSIM~\cite{wang2004image} are commonly used to assess reconstruction fidelity at the pixel level, while LPIPS~\cite{LPIPS} evaluates perceptual similarity aligned more closely with human judgments. 
In addition, FVD~\cite{FVD} effectively measures the perceptual quality and temporal diversity of video-based generative outputs, while the CLIP score~\cite{CLIP} quantifies semantic coherence between generated content and textual descriptions. 
However, these existing metrics, although widely employed, primarily address either spatial or temporal characteristics separately and do not explicitly capture critical 4D-specific challenges such as cross-view consistency and intricate spatio-temporal coherence.

Consequently, there remains a pressing need to establish comprehensive and specialized evaluation benchmarks specifically tailored to the unique demands of 4D generation tasks. 
Inspired by structured evaluation frameworks from related generative domains—such as compositional realism assessments in T2I-CompBench~\cite{huang2023t2i}, automated multimodal evaluations from DreamBench++~\cite{peng2024dreambench}, and rigorous temporal coherence metrics from video benchmarks like EvalCrafter~\cite{liu2024evalcrafter}, VBench~\cite{huang2024vbench}, and FETV~\cite{liu2023fetv}—future research should develop dedicated benchmarks specifically addressing multiview and multi-temporal consistency, nuanced semantic controllability, and multimodal alignment in dynamic 4D contexts. 
Such human-aligned and standardized benchmarks are essential for enabling objective, scalable, and reproducible model evaluations, guiding meaningful comparisons, and ultimately accelerating methodological progress within this area.

\vspace{-5pt}
\section{Conclusion} \label{sec:conclusion}
In this survey, we have presented a systematic review of recent advancements in 4D generation, a rapidly evolving and emerging research area.
We comprehensively analyze representation methods, generative model architectures, and optimization strategies, highlighting key challenges, such as consistency, controllability, diversity, efficiency, and fidelity.
Additionally, we discuss important open challenges, including dataset construction, unified model frameworks, and effective benchmarks, providing clear directions for future research and methodological innovation.
By synthesizing foundational concepts, current methodologies, and prospective opportunities, this survey serves as a comprehensive reference for new researchers and a detailed guide for practitioners.

\bibliographystyle{IEEEtran}
{
\small
\bibliography{IEEEabrv,ref}
}

\vspace{-60pt}
\begin{IEEEbiography}[{\includegraphics[width=0.8in, clip, keepaspectratio]{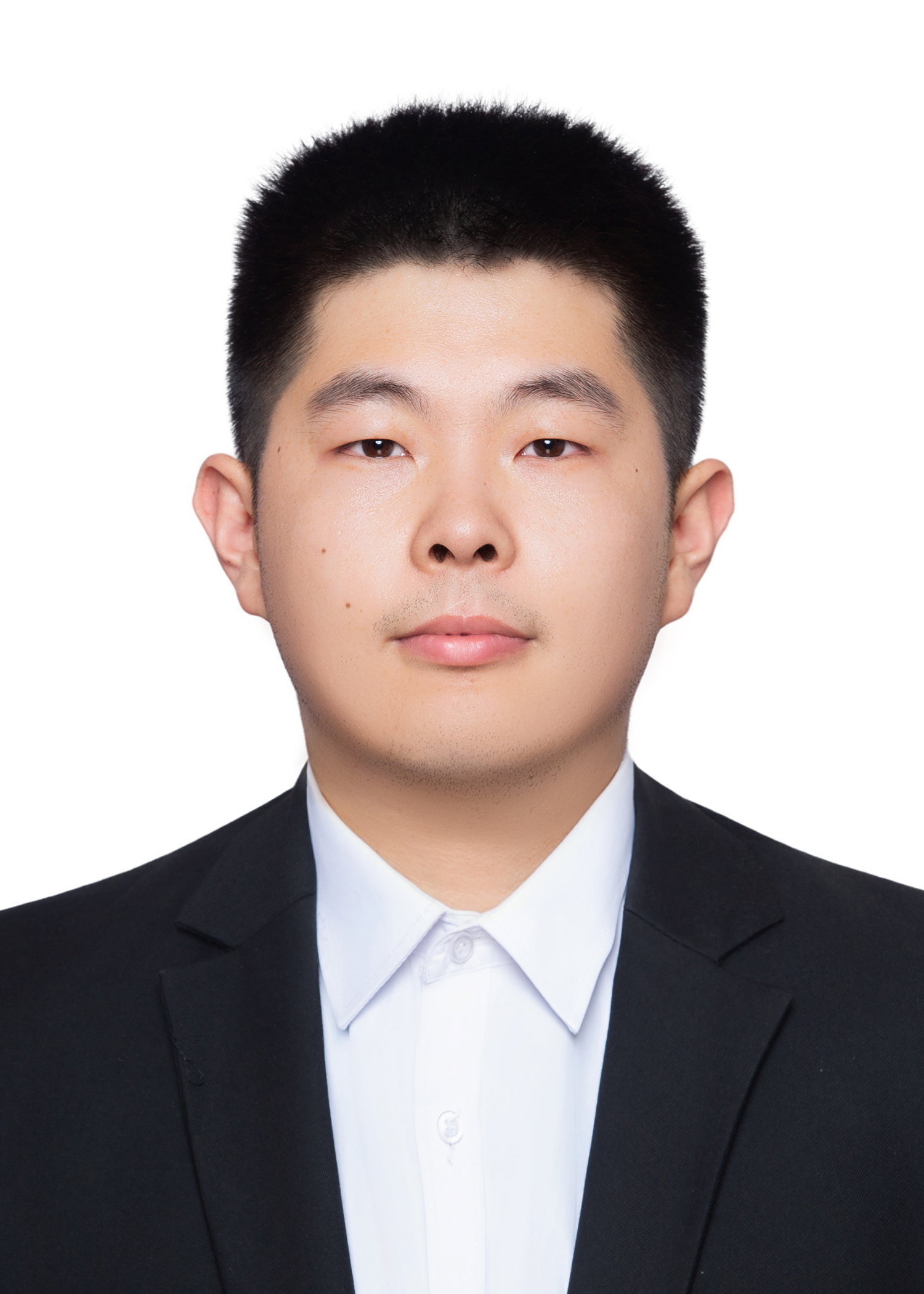}}]{Qiaowei Miao} received a bachelor’s degree in computer science and technology from Hebei University, China, in 2021. He is working
toward a PhD at the School of Software Technology, Zhejiang University, China. His research interests include 4D vision and deep learning.
\end{IEEEbiography}
\vspace{-60pt}

\begin{IEEEbiography}[{\includegraphics[width=0.8in, clip, keepaspectratio]{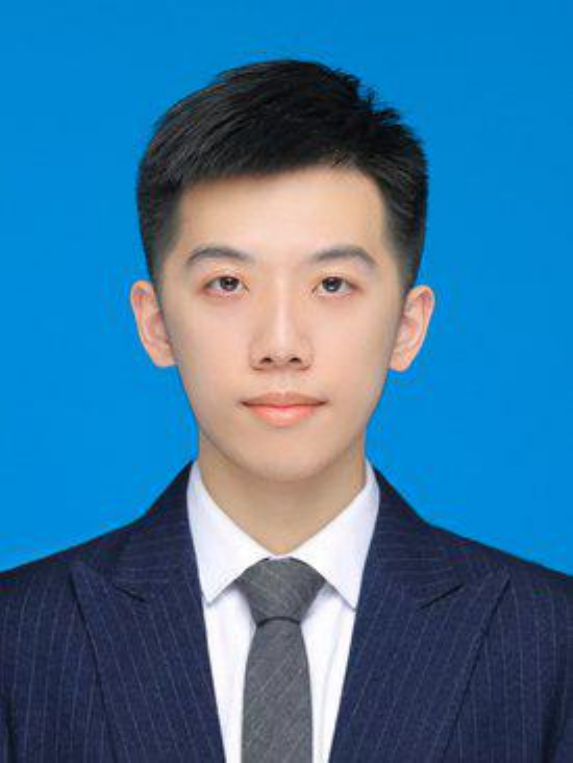}}]{Kehan Li}
received a bachelor's degree in computer science and technology at Chongqing University, China, in 2024. He is pursuing a master's degree at Zhejiang University, focusing on Artificial Intelligence and Computer Vision.
\end{IEEEbiography}
\vspace{-40pt}

\begin{IEEEbiography}[{\includegraphics[width=0.8in, clip, keepaspectratio]{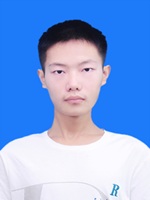}}]{Jinsheng Quan} received his bachelor’s degree in computer science and technology from Hunan Normal University. He is currently pursuing a master’s degree at Zhejiang University. His research interests include computer vision and deep learning.
\end{IEEEbiography}
\vspace{-48pt}

\begin{IEEEbiography}[{\includegraphics[width=0.8in, clip, keepaspectratio]{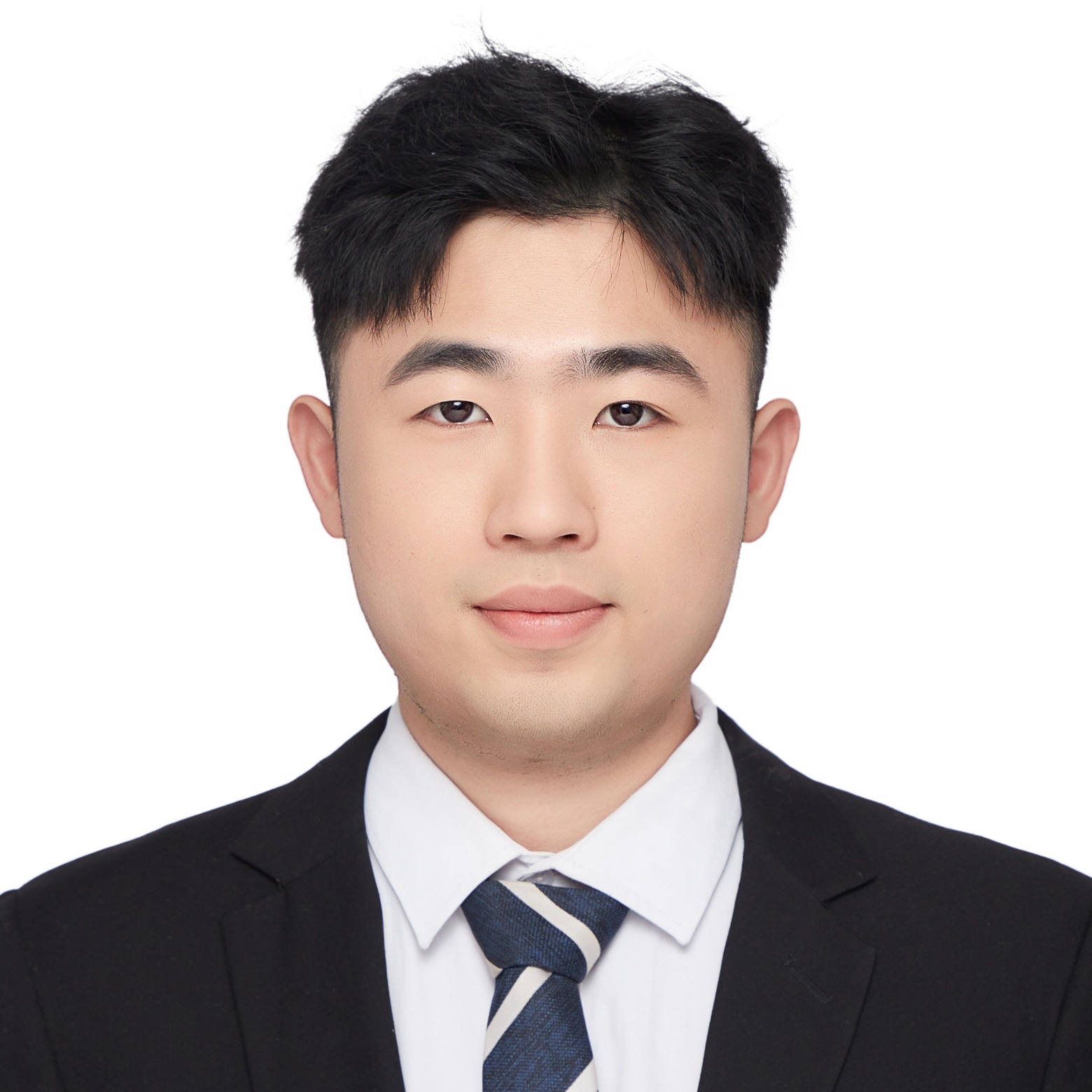}}]{Zhiyuan Min} received a Bachelor of Science degree from Wuhan University of Technology. He received a degree in Information and Computing Sciences, majoring in Computer Vision and Computer Graphics, with a focus on Artificial Intelligence. He is currently pursuing the PhD degree at the School of Computer Science, Zhejiang University.
\end{IEEEbiography}
\vspace{-48pt}

\begin{IEEEbiography}[{\includegraphics[width=0.8in, clip, keepaspectratio]{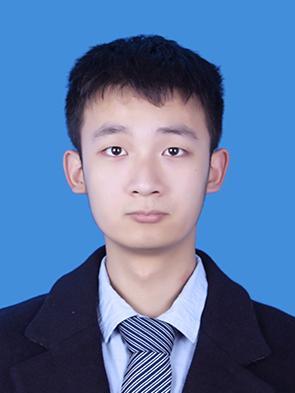}}]{Shaojie Ma} received a Bachelor's degree in Software Engineering from the School of Computer Science and Artificial Intelligence at Zhengzhou University, with a focus on Artificial Intelligence. He is currently pursuing a Master's degree at Zhejiang University.
\end{IEEEbiography}
\vspace{-48pt}

\begin{IEEEbiography}[{\includegraphics[width=0.8in, clip, keepaspectratio]{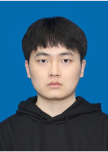}}]{Yichao Xu} received his bachelor's degree in computer science from Zhejiang University of Science and Technology. He is currently pursuing a master's degree at Zhejiang University.
\end{IEEEbiography}
\vspace{-34pt}

\begin{IEEEbiography}[{\includegraphics[width=0.8in, clip, keepaspectratio]{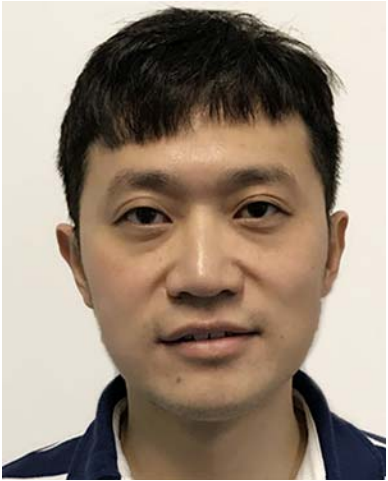}}]{Yi Yang}
received the PhD degree in computer science from Zhejiang University, Hangzhou, China,
in 2010. He is currently a distinguished professor
with Zhejiang University, China. He was a professor and director with the ReLER Lab, Australian
Artificial Intelligence Institute (AAII), University of
Technology Sydney, Australia. He was a postdoctoral research with the School of Computer Science,
Carnegie Mellon University, Pittsburgh, PA, USA.
His current research interest include machine learning
and its applications to multimedia content analysis
and computer vision, such as multimedia indexing and retrieval, surveillance
video analysis and video semantics understanding.
\end{IEEEbiography}

\vspace{-34pt}
\begin{IEEEbiography}[{\includegraphics[width=0.8in, clip, keepaspectratio]{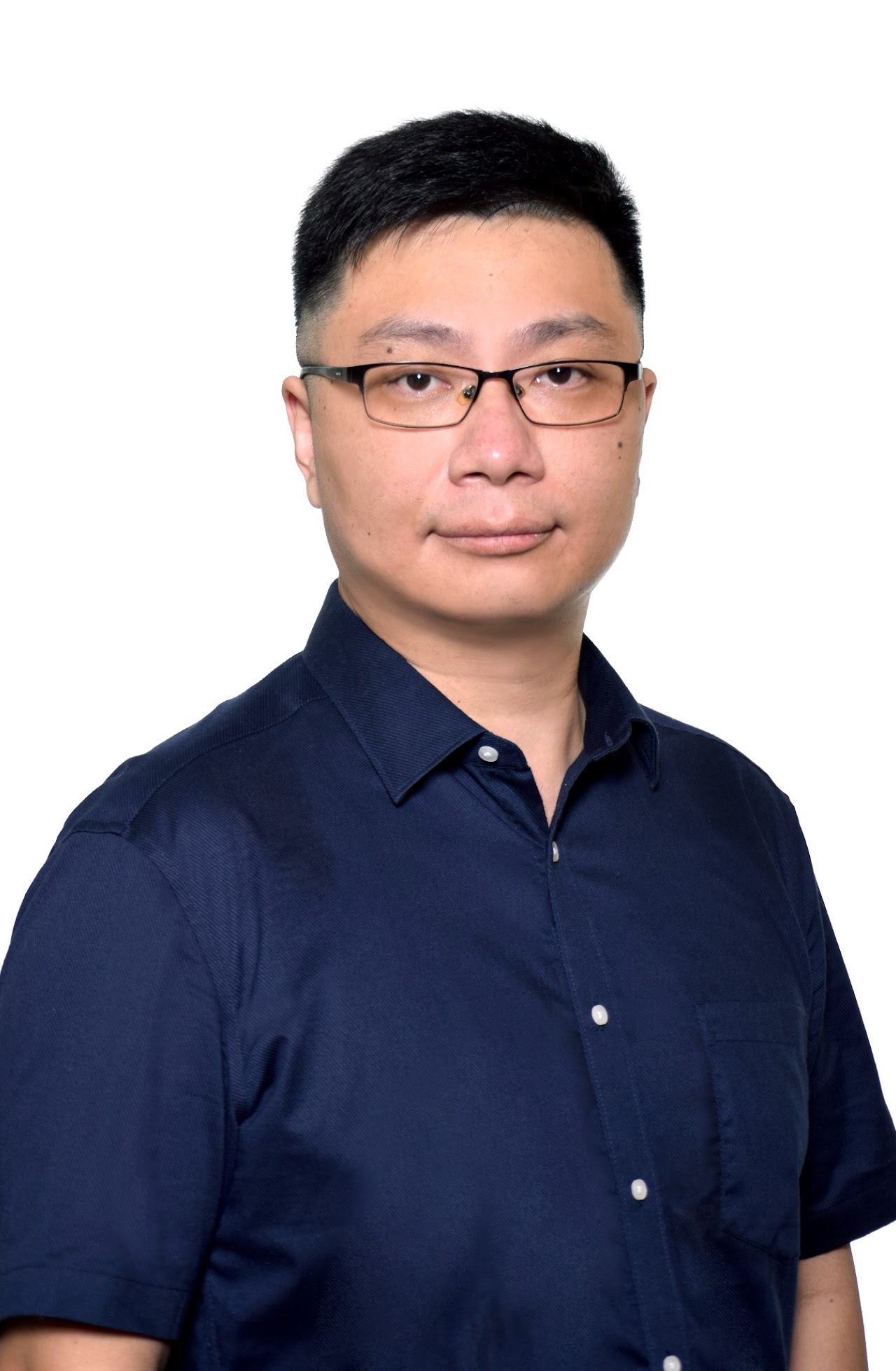}}]{Ping Liu} is an Assistant Professor in Computer Science department, University of Nevada, Reno, USA. He was a senior scientist at Centre for Frontier AI Research (CFAR), Agency for Science, Technology and Research (A*STAR), Singapore from 2020 to 2024.  From 2018 to 2020, he was a Research Staff with the Center for Artificial Intelligence, University of Technology Sydney, Ultimo, NSW, Australia.  He received the bachelor’s degree in electrical engineering from the Wuhan University of Technology, Wuhan, China, in 2005, the master’s degree from the Huazhong University of Science and Technology, Wuhan, in 2008, and the Ph.D. degree in computer science and engineering from the University of South Carolina, Columbia, SC, USA, in 2015. His research interests include computer vision and deep learning.
\end{IEEEbiography}
\vspace{-34pt}

\begin{IEEEbiography}[{\includegraphics[width=0.8in, clip, keepaspectratio]{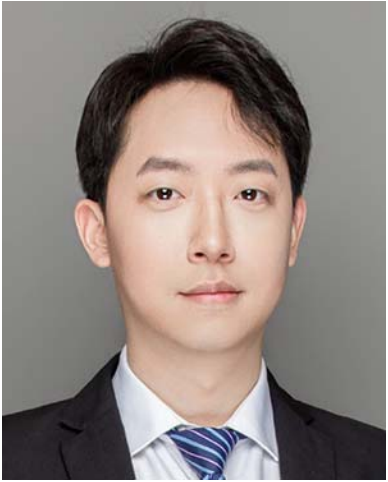}}]{Yawei  Luo}
received the PhD degree from the
Huazhong University of Science and Technology, in
2020. He is a ZJU 100 young professor with the
School of Software Technology, Zhejiang University.
He was a postdoctoral researcher with CCAI, College
of Computer Science and Technology in Zhejiang
University from 2020 to 2023. He was a visiting Ph.D
student with ReLER lab, AAII, University of Technology Sydney, from 2017 to 2019. His research interests include knowledge engineering, domain adaptation, and 3D reconstruction.
\end{IEEEbiography}

\vfill
\end{document}